\documentclass[letterpaper]{article} % DO NOT CHANGE THIS
\usepackage{prior}  % DO NOT CHANGE THIS  Anonymous submission
\usepackage[table]{xcolor} % Add this line to use \rowcolor

\usepackage{times}  % DO NOT CHANGE THIS
\usepackage{helvet}  % DO NOT CHANGE THIS
\usepackage{courier}  % DO NOT CHANGE THIS
\usepackage[hyphens]{url}  % DO NOT CHANGE THIS
\usepackage{graphicx} % DO NOT CHANGE THIS
\usepackage{natbib}  % DO NOT CHANGE THIS AND DO NOT ADD ANY OPTIONS TO IT
\usepackage{caption} % DO NOT CHANGE THIS AND DO NOT ADD ANY OPTIONS TO IT
\usepackage{algorithm}
\usepackage[T1]{fontenc}
\usepackage{epsfig}
\usepackage{amsmath}
\usepackage{amssymb}
\usepackage{booktabs}
\usepackage{bbding}
\usepackage{pifont}
\usepackage{wasysym}
\usepackage{utfsym}
\usepackage{fontawesome}
\usepackage{color}
\usepackage{algorithmicx}  
\usepackage{algpseudocode}
\usepackage{comment}
\usepackage{tablefootnote}
\usepackage{multirow}
\usepackage{subcaption}

\newcommand{\mapnet}{PriorMapNet}
\definecolor{dark-gray}{gray}{0.30}
\newcommand{\pub}[1]{{\color{dark-gray}{\scriptsize{[{#1}]}}}}

\usepackage{listings}
\DeclareCaptionStyle{ruled}{labelfont=normalfont,labelsep=colon,strut=off} % DO NOT CHANGE THIS
\floatstyle{ruled}
\newfloat{listing}{tb}{lst}{}
\floatname{listing}{Listing}
\title{{\mapnet}: Enhancing Online Vectorized HD Map Construction with Priors}
\author{
    Rongxuan Wang $^{1\dagger}$ \hspace{0.25cm} Xin Lu $^{2}$ \hspace{0.25cm} Xiaoyang Liu$^2$ \hspace{0.25cm} Xiaoyi Zou$^{2}$ \hspace{0.25cm} Tongyi Cao$^2$ \hspace{0.25cm} Ying Li$^{*1}$ \\ %  \dagger \ddagger
}
\affiliations{
    $^1$Beijing Institute of Technology \hspace{0.9cm} $^2$DeepRoute.ai \\
    rongxuan\_wang@foxmail.com \hspace{0.5cm} ying.li@bit.edu.cn

}

\usepackage{bibentry}
\begin{document}
\nocopyright

\maketitle

\let\thefootnote\relax\footnotetext{$\dagger$ Work done during the internship at DeepRoute.ai.} 
\let\thefootnote\relax\footnotetext{* Corresponding author.} % $\ddagger$ Corresponding author.

\begin{abstract}
Online vectorized High-Definition (HD) map construction is crucial for subsequent prediction and planning tasks in autonomous driving. Following MapTR paradigm, recent works have made noteworthy achievements. However, reference points are randomly initialized in mainstream methods, leading to unstable matching between predictions and ground truth. 
To address this issue, we introduce {\mapnet} to enhance online vectorized HD map construction with priors. 
We propose the PPS-Decoder, which provides reference points with position and structure priors. Fitted from the map elements in the dataset, prior reference points lower the learning difficulty and achieve stable matching. 
Furthermore, we propose the PF-Encoder to enhance the image-to-BEV transformation with BEV feature priors.
Besides, we propose the DMD cross-attention, which decouples cross-attention along multi-scale and multi-sample respectively to improve efficiency. 
Our proposed {\mapnet} achieves state-of-the-art performance in the online vectorized HD map construction task on nuScenes and Argoverse2 datasets. 
The code will be released publicly soon.
    
\end{abstract}
 \section{Introduction}
 \label{sec:intro}
High-Definition (HD) map is integral to autonomous driving, offering detailed information on critical elements such as road boundaries, traffic lanes, and pedestrian crossings~\cite{li2022hdmapnet, liao2022maptr}. This detailed information is crucial for subsequent tasks like trajectory forecasting~\cite{liang2020learning, zhou2022hivt} and path planning~\cite{hu2023planning}.
Traditionally, HD map has been constructed using offline SLAM-based methods, which are time-consuming and do not scale effectively with the rapid updates of urban environments and road networks. 
% prone to localization errors
% Furthermore, offline HD map annotation 
% , as it often fails to keep up with the pace of infrastructure changes.
To address these challenges, there is growing interest in online HD map construction methods that use vehicle-mounted sensors to generate maps in real-time. Early approaches~\cite{li2022bevformer, peng2023bevsegformer} focus on semantic segmentation in bird’s eye view~(BEV). However, these methods primarily predict rasterized maps, which lack the vectorized map information for autonomous driving tasks.

\begin{figure}[t!]
	\centering
        {\includegraphics[width=1.0\linewidth]{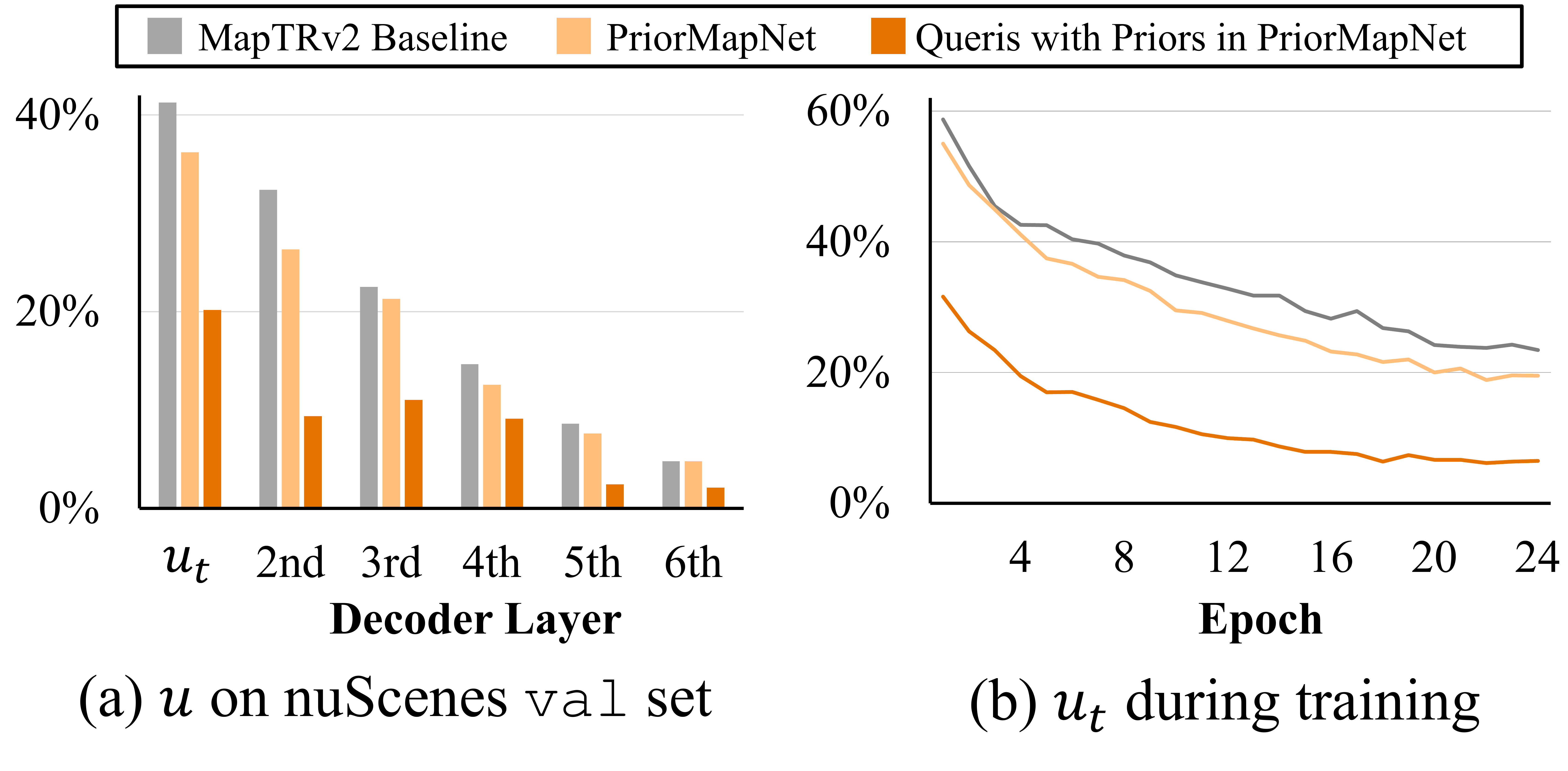}}
	\caption{Comparison of the unstable matching scores, the lower, the better. (a) and (b) denote the unstable matching scores during validation and training, respectively. $u$ means the percentage of queries whose GT match changed compared with the previous decoder layer, and $u_t$ means the percentage of final output queries whose GT match changed compared with the first decoder layer. ``Queris with Priors'' denote the queries corresponding to Prior Reference Points.}
 % Enhanced with priors, our {\mapnet} achieves more stable matching than MapTRv2.
	\label{fig:stability}
\end{figure}

Following DETR~\cite{carion2020detr} paradigm, recent advancements have introduced end-to-end learning frameworks aimed at directly predicting vectorized instances. 
MapTR~\cite{liao2022maptr} and MapTRv2~\cite{liao2023maptrv2} design an instance-point level hierarchical query embedding scheme and have demonstrated promising outcomes in constructing vectorized HD maps. 
% MapTR series methods consist of three modules: an image feature extraction backbone, an image-to-BEV feature transformation encoder, and a Transformer-based map element detection decoder. In the Transformer decoder, map elements are represented by learnable queries, which are optimized layer by layer. 
The mainstream methods proposed later follow this pipeline, with improvements focusing on enhancing interactions between queries and integrating external features~\cite{xu2023InsMapper, liu2024mgmap, zhou2024himap}.
% with improvements focusing on interactions between query internals or with external features. 

In these methods, queries learn the position and structure of map elements and are matched with the ground truth~(GT) during training. However, the Hungarian algorithm used for matching is sensitive to small changes in the cost matrix, which leads to \textit{unstable matching}~\cite{li2022dn}. To quantify the unstability of matching, we define the unstable matching score $u$ following Stable-DINO~\cite{liu2023stabledino}, representing the percentage of queries whose GT match changed compared with the previous decoder layer. We also measure the total unstable matching score $u_t$, which represents the percentage of final output queries whose GT match changed compared with the first decoder layer. As shown in Fig.~\ref{fig:stability}, MapTRv2 exhibits unstable matching throughout the process of training and validation.
%, which results in reduced accuracy.

\begin{figure}[t!]
\centering
\includegraphics[scale=0.095]{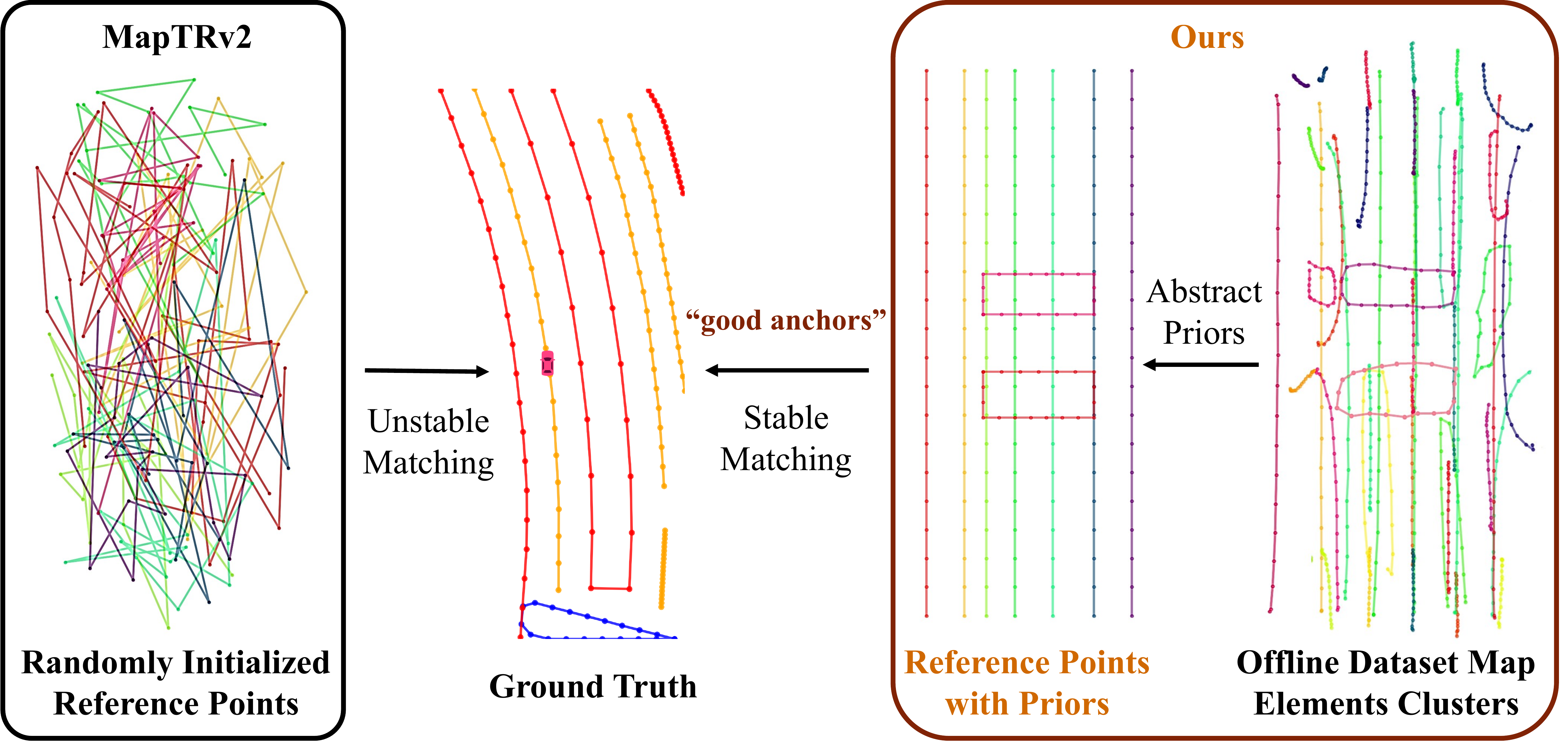}
\caption{Comparison of the matching of MapTRv2 and our proposed method. Reference points with position and structure priors achieve stable matching.}
\label{fig:matching}
\end{figure}

Why the matching is unstable? The training process of DETR-like models has two stages: learning ``good anchors''~(stage I) and learning relative offsets~(stage II)~\cite{li2022dn}. In mainstream methods, queries consist of content embeddings and position embeddings. Position embeddings generate reference points for sampling~(related to stage I), and content embeddings generate sampling offsets and attention weights~(related to stage II). 
Position embeddings are learnable and initialized randomly, which leads to reference points that are distributed without any specific structure. In contrast, the vectorized HD map consists of map elements like polylines or polygons connected in an ordered sequence, with distinct positional distributions and geometric patterns. As shown in Fig.~\ref{fig:matching}, matching these structured map elements with randomly distributed reference points is challenging and results in unstable matching.
% Position embeddings are learnable and randomly initialized, resulting in randomly distributed reference points. However, for vectorized HD map, map elements are polylines or polygons connected by an ordered point sequence, and their positional distribution and geometric structure have certain patterns. Letting the randomly initialized reference points match the structured map elements is inherently difficult, as illustrated in Fig.~\ref{fig:matching}. It is the randomly initialized reference points that lead to unstable matching results and reduced accuracy. 
To solve this problem, we propose the Decoder with Prior Position and Structure~(PPS-Decoder). By fitting the distribution of map elements in datasets through clustering and abstracting these distributions as priors, the reference points are enhanced to better match the positional and structural features of map elements. As demonstrated in Tab.~\ref{tab:ablation-main}, prior-aware queries improve both accuracy and matching stability by reducing the difficulty of learning ``good anchor''.

In essence, the prior is an effective initialization method, reducing the learning difficulty for the model. To leverage this approach, we introduce the Encoder with Prior Feature (PF-Encoder). PF-Encoder transforms image features into initialized BEV features, which are utilized as BEV query priors and optimized in the encoder. Discriminative Loss is introduced to better aggregate map elements embeddings. Besides, BEV features are downsampled to multi-scale, bringing computational complexity. 
% which uses LSS~\cite{philion2020lift} to transform image features into initialized BEV feature. The BEV features are used as BEV query priors, optimized in a BEVFormer~\cite{li2022bevformer} encoder, and downsampled to multi-scale. 
To enhance efficiency, we propose the Decoupled Multi-Scale Deformable Cross-Attention~(DMD cross-attention), which decouples cross-attention along multi-scale and multi-sample respectively. The combination of the PF-Encoder, PPS-Decoder, and DMD cross-attention forms our proposed {\mapnet}.

% evaluation
Extensive experiments are conducted to prove our superiority. We achieve state-of-the-art (SOTA) performance in online vectorized HD map construction on nuScenes~\cite{caesar2020nuscenes} and Argoverse2~\cite{wilson2023argoverse} datasets. Furthermore, experiments conducted under various settings demonstrate the robustness and generalization capabilities of {\mapnet}.
In summary, our contributions are:

\begin{itemize}
\item We introduce a novel prior-based framework for online HD map construction by integrating feature, position, and structure priors into encoder and decoder.
\item We propose the DMD cross-attention, which decouples cross-attention along multi-scale and multi-sample respectively to improve efficiency.
\item We achieve SOTA performance in online vectorized HD map construction on the nuScenes and Argoverse2 datasets, demonstrating both high performance and generalization capability.
\end{itemize}

\section{Related Work}

\subsection{Online Vectorized HD Map Construction}
% HD map plays a crucial role in autonomous driving, aiming to provide detailed information about map elements. 
Unlike traditional offline HD map construction methods, recent studies use vehicle-mounted sensors to construct online HD map. Early methods~\cite{philion2020lift, li2022bevformer, liu2022petrv2} tackle map construction as a segmentation task, predicting rasterized maps in BEV space. HDMapNet~\cite{li2022hdmapnet} further converts these rasterized maps into vectorized maps through post-processing.

VectorMapNet~\cite{liu2023vectormapnet} introduces the first end-to-end vectorized map model, using a DETR~\cite{carion2020detr} decoder to detect map elements and optimizing results with an auto-regressive transformer. 
Subsequently, MapTR~\cite{liao2022maptr} and MapTRv2~\cite{liao2023maptrv2} design a one-stage map construction paradigm with an instance-point level hierarchical query embedding scheme. 
The mainstream methods proposed later follow this pipeline, with improvements focusing on enhancing interactions of queries and external features. 
InsMapper~\cite{xu2023InsMapper} and HIMap~\cite{zhou2024himap} further explore the correlation between instances and points and improve the interaction within queries. MapQR~\cite{liu2024mapqr} implicitly encodes point-level queries within instance-level queries and embeds query positions like Conditional DETR~\cite{meng2021conditional}. Despite the above developments, these methods randomly initialize reference points, resulting in unstable matching. To address this issue, our {\mapnet} introduces priors to enhance matching stability.

% MapTR series methods consist of three modules: an image feature extraction backbone, an image-to-BEV feature transformation encoder, and a Transformer-based map element detection decoder. In the Transformer decoder, map elements are represented by learnable queries, which are optimized layer by layer. 

\begin{figure*}[t]
\centering
\includegraphics[scale=0.12]{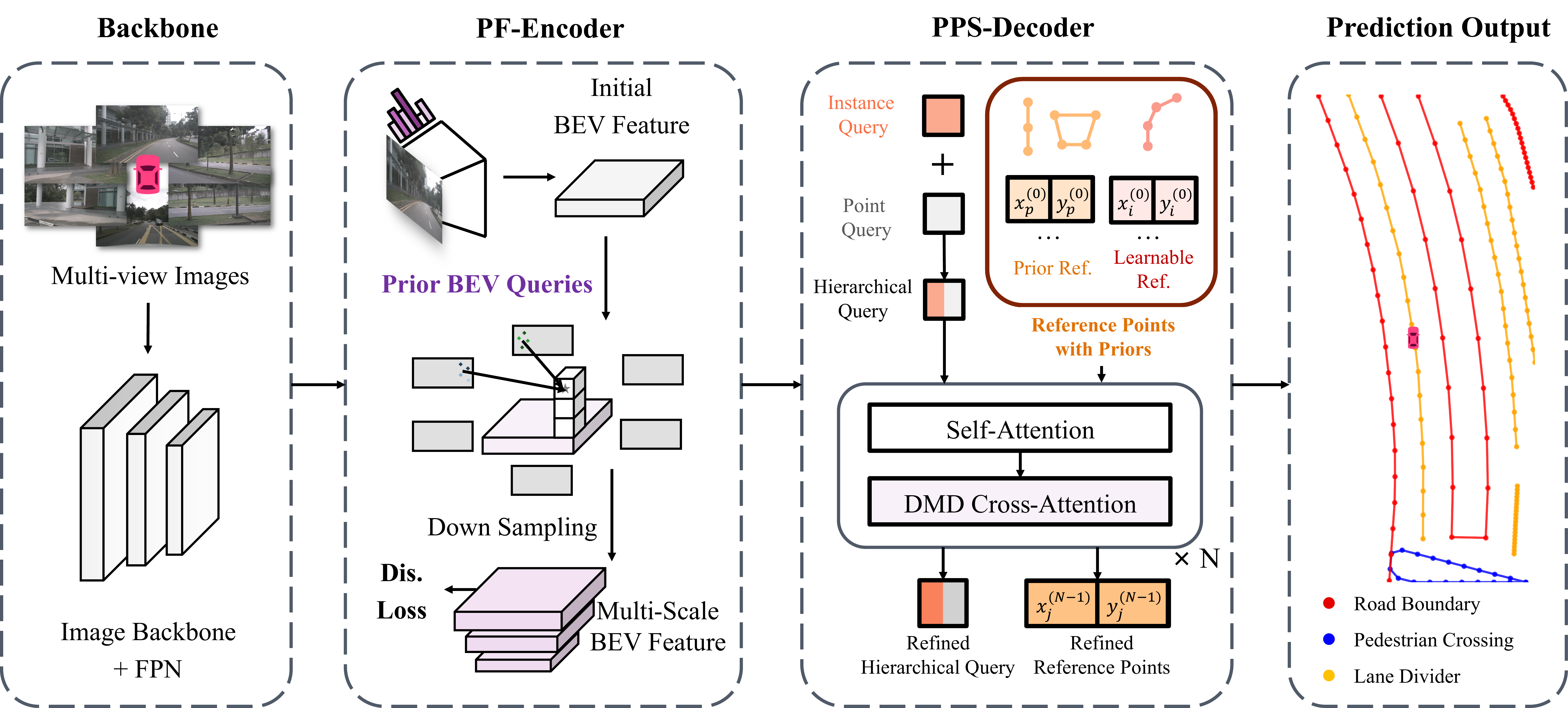}
\caption{The overview of our proposed {\mapnet}. 
Given multi-view images as input, the output is a set of map elements. {\mapnet} consists of three modules: the backbone, the PF-Encoder and the PPS-Decoder. The backbone extracts image features by using the ResNet and a FPN neck. The PF-Encoder transforms image features into BEV and downsamples it to multiple scales. The PPS-Decoder predicts map elements through Transformer, and reference points with priors are used for stable matching. In the cross-attention layer, the DMD cross-attention is used to achieve efficiency.
}
\label{fig:pipeline}
\end{figure*}

\begin{figure*}[t]
	\centering
        \subfloat[MapTRv2 Decoder
    	\label{fig:decoder-maptrv2}
        ]{
	\begin{minipage}{0.31\linewidth}{
			\begin{center}
            {\includegraphics[width=1\linewidth]{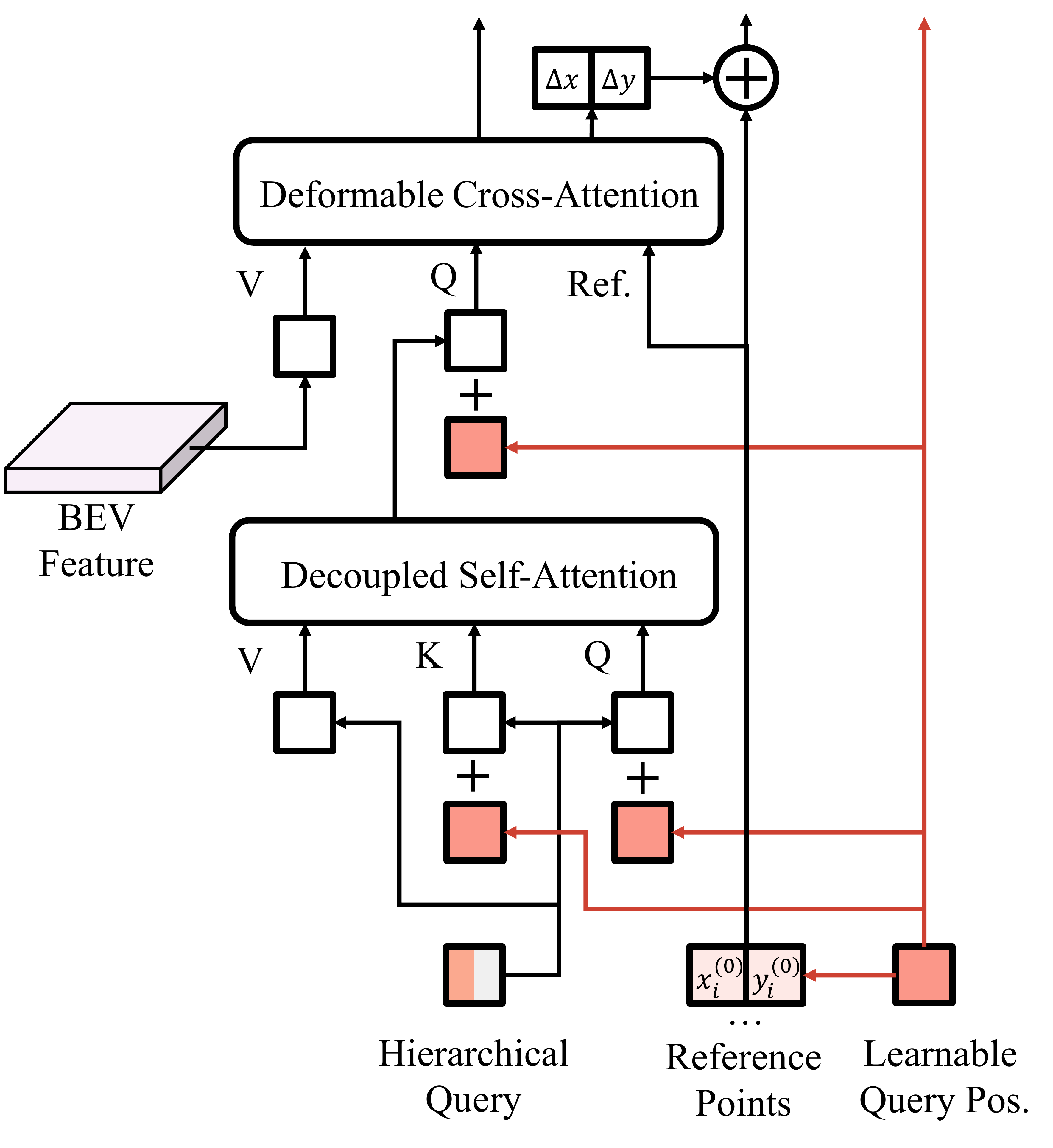}}
			\end{center}
		}
	\end{minipage}
        }
        \subfloat[MGMap Decoder
    	\label{fig:decoder-mgmap}
        ]{
	\begin{minipage}{0.32\linewidth}{
			\begin{center}
            {\includegraphics[width=1\linewidth]{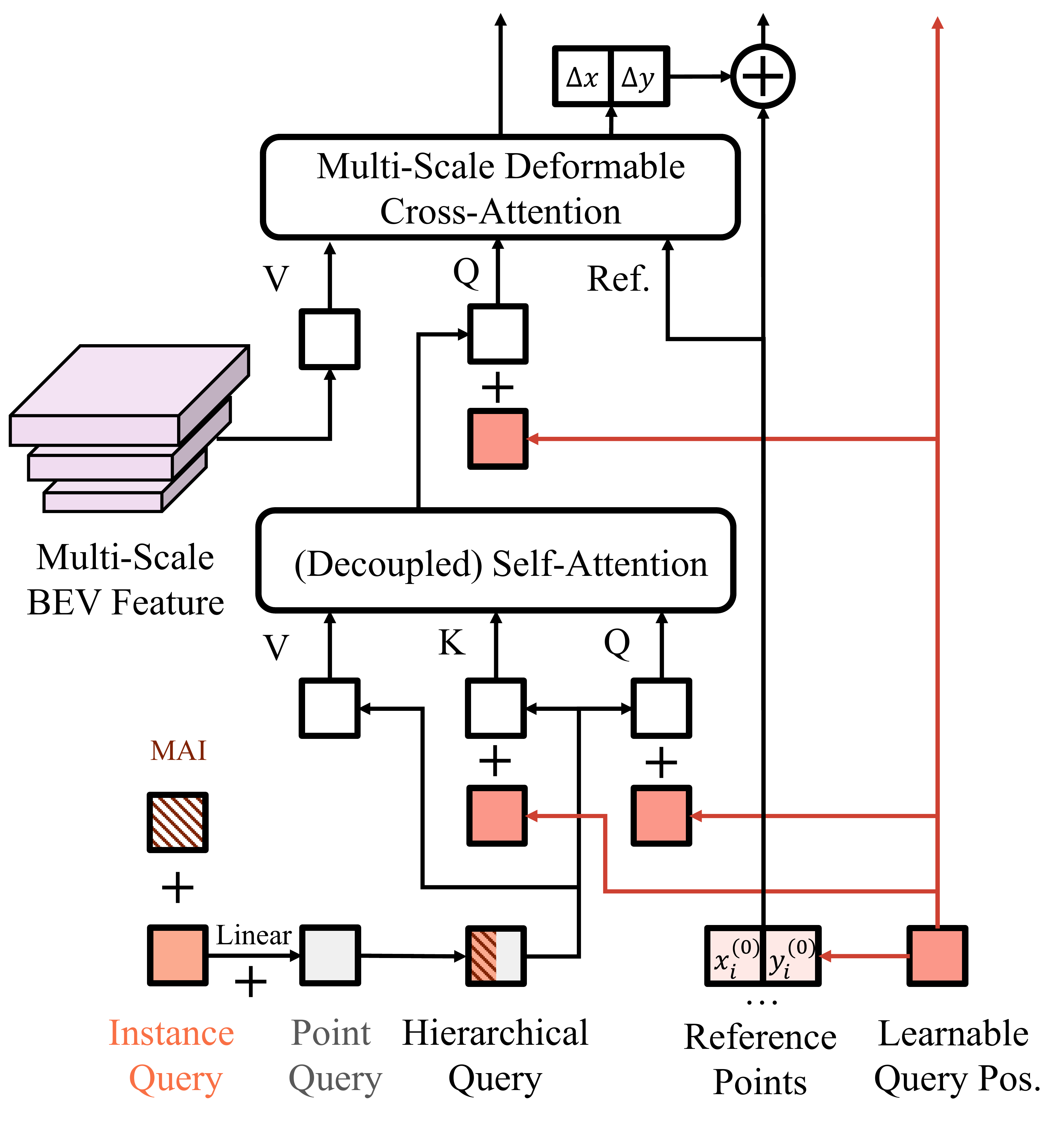}}
			\end{center}
		}
	\end{minipage}
        }
        \subfloat[Our PPS-Decoder
            \label{fig:decoder-proposed}
        ]{
        \begin{minipage}{0.32\linewidth}{
			\begin{center}
			{\includegraphics[width=1\linewidth]{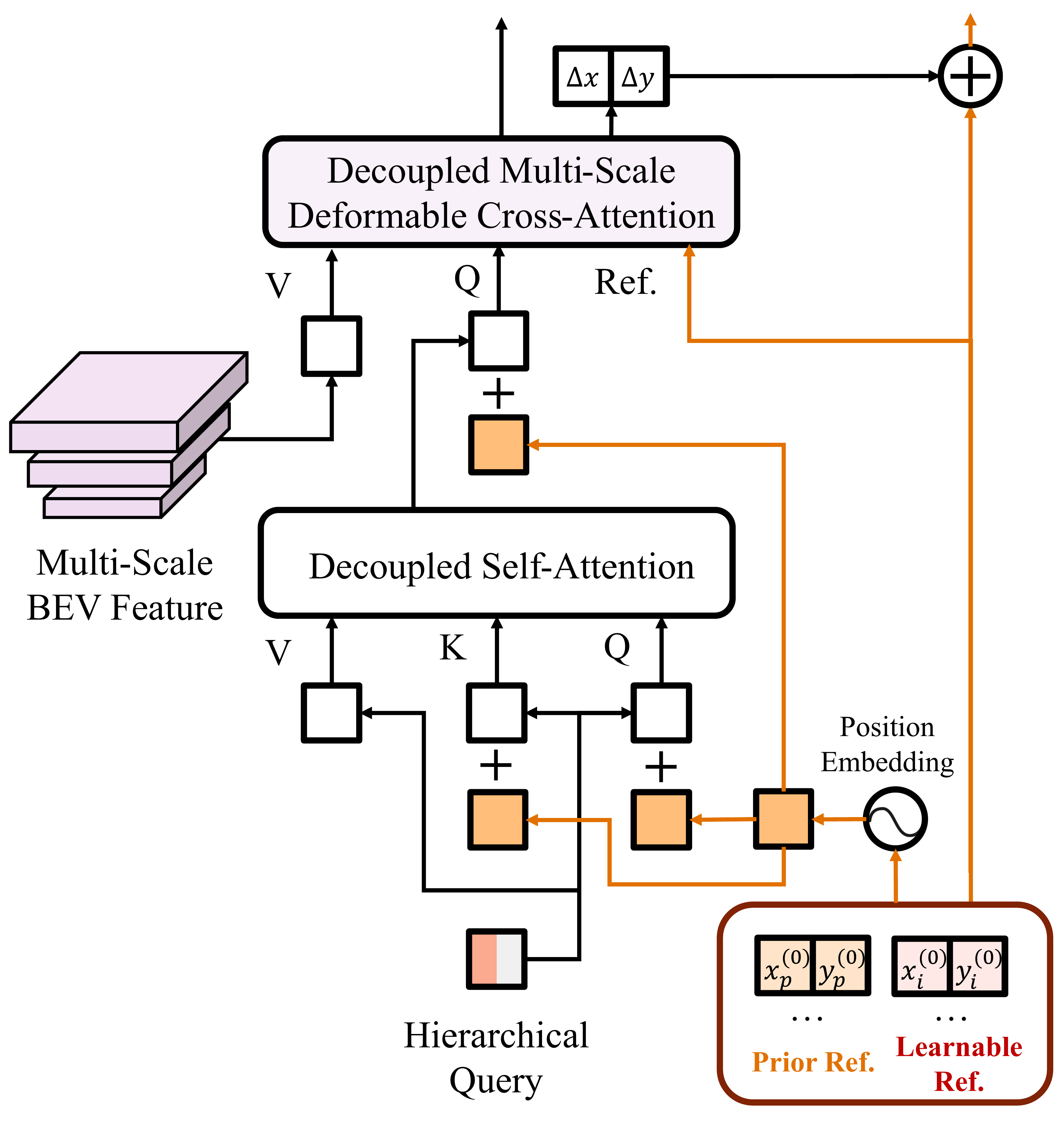}}
			\end{center}
		}
	\end{minipage}
        }
	\caption{Comparison of the decoder of MapTRv2, MGMap and our proposed {\mapnet}. For simplicity, we only show the first layer in the transformer decoder. (a) MapTRv2 uses randomly initialized learnable query positions for all layers without any adaptation, which brings unstable matching results. (b) MGMap adds Mask-Activated Instance to provide semantic priors, but lacks position information. In contrast, (c) {\mapnet} enhances reference points with priors, which achieves stable matching.}
	\label{fig:decoder}
\end{figure*}

\subsection{Priors for HD Map Construction}
Priors provide effective initialization for map construction and reduce the difficulty of model learning. We categorize priors into two types: semantic priors and positional and structural priors. For prior semantics, MGMap~\cite{liu2024mgmap} proposes Mask-Active Instance~(MAI), which learns map instance segmentation results and provides semantic priors for instance queries. Bi-Mapper~\cite{li2023bimapper} designs a two-stream model, using priors from global and local perspectives to enhance semantic map learning. For prior position and structure, Topo2D~\cite{li2024topo2d} uses 2D lane detection results as priors to initialize queries. SMERF~\cite{luo2023smerf} and P-MapNet~\cite{jiang2024pmapnet} introduce Standard Map~(SDMap) as position and structure priors for map construction. However, the above methods rely on additional modules, increasing computational complexity. In contrast, {\mapnet} uses offline clustered map elements as position and structure priors,  improving performance without additional computational consumption.
% Prior position and structure are more important in vectorized map construction because position is the core of query design and regression.

\subsection{Image-to-BEV Encoder for Map Construction}
Map construction usually relies on the BEV feature, which is transformed from images by the encoder. There are two types of encoders: bottom-up and top-down. Bottom-up encoders~\cite{philion2020lift, huang2021bevdet, li2022bevstereo, li2023bevdepth} lift images to 3D and use voxel pooling to generate BEV features. Top-down encoders~\cite{wang2022detr3d, li2022bevformer, yang2023bevformer, chen2022gkt} generate BEV queries containing 3D information and extract image features to BEV queries with the transformer. However, since queries are randomly initialized, the single-layer encoder results in low accuracy~\cite{liao2022maptr}, and the multi-layer encoder brings more computational complexity~\cite{liu2024mapqr, lilanesegnet}. To overcome these limitations, we enhance BEV queries with prior features.

\section{Method}

\subsection{Overview}
Fig.~\ref{fig:pipeline} shows the overall pipeline of our method. Given $N_c$ multi-view images $\{{ {I}_i}\}_{i=1}^{N_c}$ as input, the output is a set of $N_m$ map elements $\{{ {M}_i}\}_{i=1}^{N_m}$. Each map element is defined as a class label $c$ and an ordered point sequence $\{{(x_i,y_i)}\}_{i=1}^{N_p}$,  where ${N_p}$ is the number of points in each map element. 

Based on MapTRv2~\cite{liao2023maptrv2}, our method consists of three modules: the backbone, the PF-Encoder and the PPS-Decoder. The backbone extracts multi-scale images features $\{{ {F}_{\mathrm{img}}^i}\}_{i=1}^{N_c}$ by ResNet~\cite{he2016resnet} and a FPN~\cite{lin2017feature} neck. The PF-Encoder transforms images features to BEV features $ {F}_{\mathrm{BEV}}\in\mathbb{R}^{{H_{{\mathrm{}}}}\times{W_{{\mathrm{}}}}\times{C}}$ and downsamples it to multiple scales, as described in Section \ref{sec:encoder}. The PPS-Decoder predicts map elements through transformer, and reference points with priors are used for stable matching, as detailed in Section \ref{sec:decoder}. In the cross-attention layer, we introduce the DMD cross-attention to achieve efficiency, as described in Section \ref{sec:crossatt}. We start by detailing the PPS-Decoder, which is the core of our method.

\subsection{Decoder with Prior Position and Structure} \label{sec:decoder}

The pipeline of our PPS-Decoder is shown in Fig.~\ref{fig:decoder-proposed}. Compared with MapTRv2, which randomly initializes reference points, and MGMap, which only provides semantic priors without position information, the PPS-Decoder enhances reference points with position and structure priors, providing ``good anchor'' to improve accuracy and matching stability.

The PPS-Decoder contains several cascaded decoder layers to refine the hierarchical queries and reference points iteratively. 
Hierarchical queries consist of instance-level queries ${q}_\mathrm{ins}\in\mathbb{R}^{N_{I}\times{C}}$ and point-level queries ${q}_\mathrm{pts}\in\mathbb{R}^{N_{P}\times{C}}$, which are combined through broadcasting:
\begin{equation}
 {q}={q}_\mathrm{ins} + {q}_\mathrm{pts},~{q}\in\mathbb{R}^{{N_{I}}\times{N_{P}\times{C}}} .
\end{equation}

% Hierarchical queries and reference points are
Reference points are initialized with prior position and structure. To fit the distribution of map elements in the dataset, we use K-Means to cluster map elements and abstract the position information of the first $N_\mathrm{pri}$ elements, as shown in Fig.~\ref{fig:matching}. Clustering and abstraction are done offline, ensuring no additional computational burden during inference. During training and inference, some reference points obtain the fitted position and structure priors (called Prior Reference Points, $R_\mathrm{pri}\in\mathbb{R}^{{N_\mathrm{pri}}\times{N_{P}}\times{2}}$), while the rest of the reference points are still from learnable parameters (called Learnable Reference Points, $R_\mathrm{lrn}\in\mathbb{R}^{{N_\mathrm{lrn}}\times{N_{P}}\times{2}}$). The combined set of reference points is denoted as $R=\left \{ R_\mathrm{pri},R_\mathrm{lrn} \right \}$, where the total number of instance queries is $N_{I}=N_\mathrm{pri}+N_\mathrm{lrn}$.

% A hierarchical query embedding scheme is used to encode each map element explicitly. 

% $P_{i,j}=\left ( x_i, y_i \right )_j$

% \begin{equation}
%  R_\mathrm{pri}=\left \{ \left \{ P_{i,j}^{\mathrm{pri}}  \right \}_{i=1}^{N_P}  \right \} _{j=1}^{N_\mathrm{pri}},~
%  R_\mathrm{lrn}=\left \{ \left \{ P_{i,j}^{\mathrm{lrn}}  \right \}_{i=1}^{N_P}  \right \} _{j=1}^{N_\mathrm{lrn}}.
% \end{equation}

% \begin{equation}
%  {q}_\mathrm{pos}=\mathrm{MLP}(\mathrm{SineEmbed}(R)).
% \end{equation}

To embed query position, reference points are encoded with sinusoidal positions following DAB-DETR~\cite{liu2022dabdetr}. Query position embedding is achieved as follows:
\begin{equation}
 {q}_\mathrm{pos}=\mathrm{Linear}(\mathrm{PE}(R)),~\mathbb{R}^{{N_{I}}\times{N_{P}}\times{2}}\to\mathbb{R}^{{N_{I}}\times{N_{P}}\times{C}},
\end{equation}
where $\mathrm{PE}(\cdot)$ generates sinusoidal embeddings based on reference points coordinates~\cite{vaswani2017attention}. The parameters of linear layers are not shared across decoder layers. $\mathrm{PE}(\cdot)$ is calculated separately on coordinates, and position embeddings are concatenated along feature channels:
\begin{equation}
 \mathrm{PE}(R)=\mathrm{PE}(x_r, y_r)=\mathrm{Cat}(\mathrm{PE}(x_r),\mathrm{PE}(y_r)).
\end{equation}

Reference points and position embeddings are updated across the PPS-Decoder layers. In each layer, self-attention and cross-attention mechanisms use the following inputs for queries, keys, values, and reference points:
\begin{equation}
\begin{cases}\vspace{1mm}
 \mathrm{Self}\text {-}\mathrm{Attn:}~\mathrm{Q}=q+{q}_\mathrm{pos},~\mathrm{K}=q+{q}_\mathrm{pos},~\mathrm{V}=q, \\
 \mathrm{Cross}\text {-}\mathrm{Attn:}~\mathrm{Q}=q+{q}_\mathrm{pos},~\mathrm{V}= {F}_{\mathrm{BEV}},~\mathrm{R}=R.
\end{cases}
\end{equation}

The Prior Reference Points fit the position and structure distribution of the map elements in the dataset, which helps queries concentrate on learning the offsets from reference points. In addition, we maintain Learnable Reference Points to capture and represent map elements that deviate from typical position and structure patterns. The self-attention enables interaction between Prior Reference Points and Learnable Reference Points, reducing redundant detections and improving overall detection accuracy.

% \begin{equation}
%  \mathrm{Cross}\text {-}\mathrm{Attn:}~\mathrm{Q}=q+{q}_\mathrm{pos},~\mathrm{V}= {F}_{\mathrm{BEV}},~\mathrm{R}=R.
% \end{equation}

\subsection{Encoder with Prior Feature} \label{sec:encoder}
PF-Encoder enhances the image-to-BEV transformation with BEV feature priors. Built on the foundation of top-down encoders, such as BEVFormer~\cite{li2022bevformer} and GKT~\cite{chen2022gkt}, PF-Encoder leverage BEV features as queries to extract relevant image features via cross-attentions.
% However, since queries are randomly initialized, the single-layer encoder has low accuracy~\cite{liao2022maptr}, and the multi-layer encoder consumes more time~\cite{liu2024mapqr, lilanesegnet}. To overcome these limitations, we enhance BEV queries with prior features.

We first utilize LSS~\cite{philion2020lift} to transform image features into initialized BEV features, which are then used as BEV query priors, optimized in a single-layer BEVFormer~\cite{li2022bevformer} encoder. Following MGMap~\cite{liu2024mgmap}, BEV features are downsampled to multi-scale with an EML neck. 

For queries to better aggregate features from the same map element, it is necessary to assimilate the embeddings of the same instance and distinguish the embeddings of different instances. Therefore, we introduce Discriminative Loss of map elements~\cite{neven2018lanenet} to bring the same instance closer and separate different instances further:

\begin{equation}
\begin{cases}\vspace{2mm} L_\mathrm{var}=\frac1{K} \sum_{k=1}^{{K}}\frac1{p_n}\sum_{l=1}^{p_n}\left[\left\|\mu_k-e_l\right\|-\delta_\mathrm{v}\right]_+^2,
\\ 
L_\mathrm{dist}=\frac1{{K}({K}-1)}\sum_{i=1}^{K}\sum_{j=1,_{i\neq j}}^{K}\left[\delta_\mathrm{d}-\left\|\mu_{i}-\mu_{j}\right\|\right]_+^2,
\end{cases}
\end{equation}
where $L_\mathrm{var}$ pulls the embeddings $e_i$ of ${K}$ map elements toward their respective means $\mu_n$, and $L_\mathrm{dist}$ pushes away the mean embeddings of different map elements. ${p_n}$ represents the number of grids of map elements. $\left \| \cdot  \right \| $ is the L2 distance and $\left [  x\right ] _+=\mathrm{max}(0,x)$. $\delta_\mathrm{v}$ and $\delta_\mathrm{d}$ are the borders of the variance and distance loss. The total Discriminative Loss is defined as $L_\mathrm{dis}=\lambda_1 L_\mathrm{var}+\lambda_2 L_\mathrm{dist}$.
% 其中$L_\mathrm{var}$将$N$个地图元素的嵌入$e_i$拉向各自的均值$\mu_n$，$L_\mathrm{dist}$推远不同元素的平均嵌入。

In the cross-attention layer of the PPS-Decoder, queries weighted sample BEV features. PF-Encoder enables the queries to effectively aggregate features associated with the same map element while distinguishing between different map instances, improving the accuracy of map construction.

\subsection{Decoupled Multi-Scale Deformable Attention} \label{sec:crossatt}
\begin{figure}[t!]
	\centering
        \subfloat[Vanilla MSDA
    	\label{fig:vanilla-msda}
        ]{
	\begin{minipage}{0.3\linewidth}{
			\begin{center}
            {\includegraphics[width=1\linewidth]{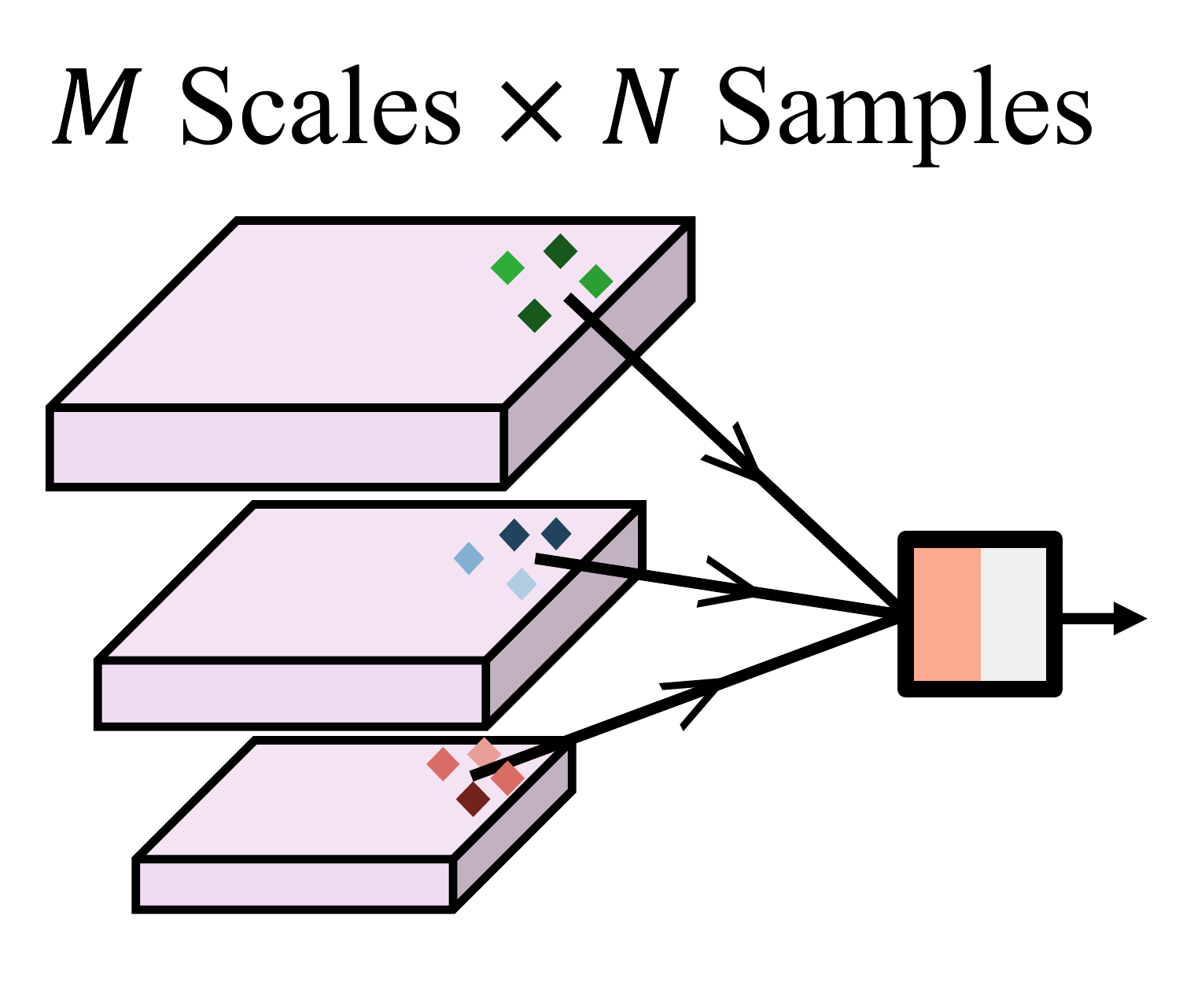}}
			\end{center}
		}
	\end{minipage}
        }
        \hspace{0.1mm}
        \subfloat[Our DMD cross-attention
            \label{fig:decoupled-msda}
        ]{
        \begin{minipage}{0.63\linewidth}{
			\begin{center}
			{\includegraphics[width=1\linewidth]{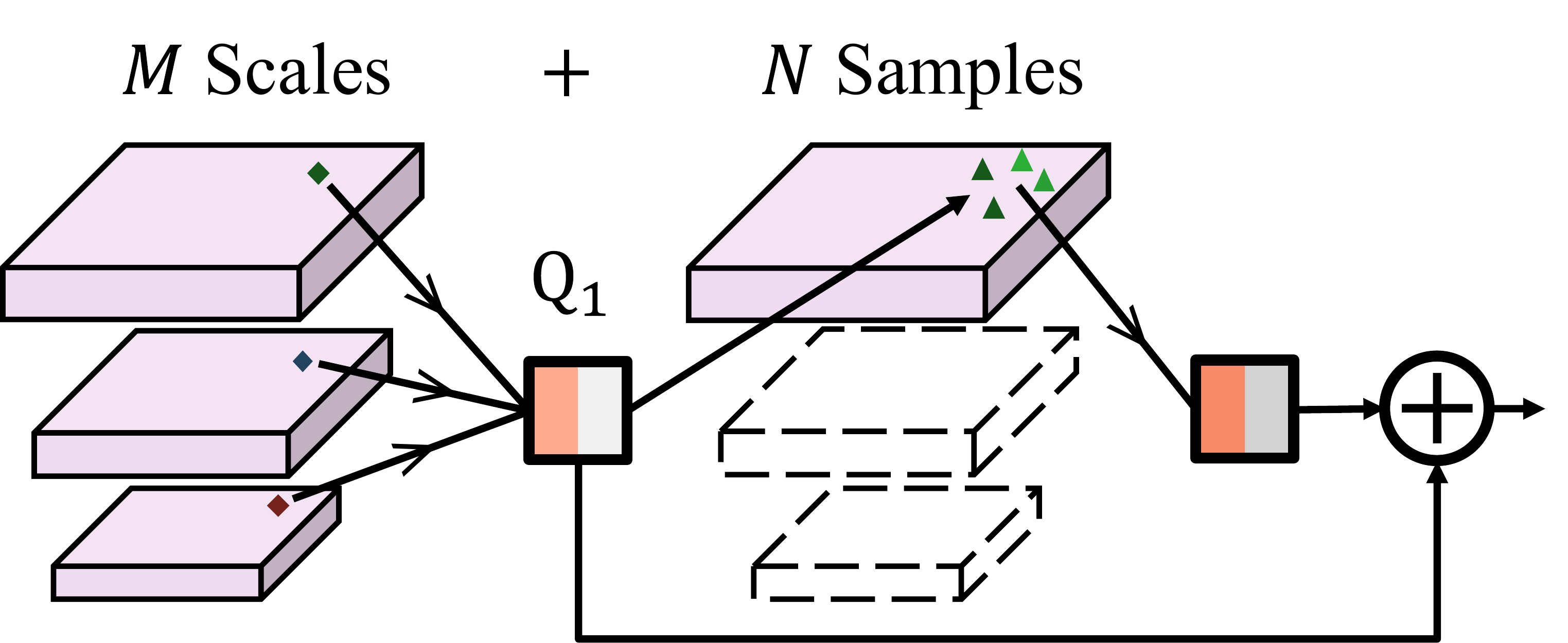}}
			\end{center}
		}
	\end{minipage}
        }
	\caption{Comparison of the vanilla MSDA and our proposed DMD cross-attention. DMD cross-attention performs cross-attention along multi-scale and multi-sample respectively to achieve efficiency.}
	\label{fig:msda}
\end{figure}

To address the computational complexity of multi-scale deformable cross-attention (MSDA), we propose the DMD cross-attention mechanism to decouple cross-attention along multi-scale and multi-sample, as shown in Fig.~\ref{fig:decoupled-msda}.

In vanilla MSDA~\cite{zhu2020deformable}, each query interacts with $M$-scale BEV features and $N$ points are sampled at each scale, whose computation complexity is $O(M\times N)$:
\begin{equation}
\mathrm{MSDA}(\mathrm{Q},\mathrm{V},\mathrm{R})=\sum_{i=1}^{N_\mathrm{h}}W_i\sum_{j=1}^{M}\sum_{k=1}^{N}A_{ijk} \cdot W^{'}_i\mathrm{V}_j(\mathrm{R}+O_{ijk}  ) ,
\end{equation}
where ${N_\mathrm{h}}$ is the number of attention heads. $A_{ijk}\in[0,1]$ and $O_{ijk}\in\mathbb{R}^2$ are attention weight and sampling offset respectively, which are generated from $\mathrm{Q}$. $A_{ijk}\in[0,1]$ is normalized by $\sum_{j=1}^{M}\sum_{k=1}^{N}A_{ijk}=1$. $W_i\in\mathbb{R}^{C \times (C/N_\mathrm{h})}$ and $W^{'}_i\in\mathbb{R}^{(C/N_\mathrm{h}) \times C}$ are learnable weights. 

To improve efficiency, the DMD cross-attention mechanism decouples the vanilla MSDA process into two stages:
% 在原始的MSDA中，每个query在M个尺度的BEV特征分别采样N个点。为了减少计算和记忆的消耗，我们引入了Decoupled Multi-Scale Deformable cross-attention，as shown in Fig.~\ref{fig:decoupled-msda}. 极大地将计算复杂度减少至$O(M+N)$.
\begin{equation}
\begin{cases}\vspace{1mm}
\mathrm{Q_1}=\mathrm{Linear}_1(\mathrm{MSDA}_1(\mathrm{Q},\mathrm{V},\mathrm{R})),
\\
\mathrm{Q_{output}} = \mathrm{Q_1}+\mathrm{Linear}_2(\mathrm{MSDA}_2(\mathrm{Q_1},{\mathrm{V}}_1,\mathrm{R})),&
\end{cases}
\end{equation}
where $\mathrm{MSDA}_1(\cdot)$ and $\mathrm{MSDA}_2(\cdot)$ denote $\mathrm{MSDA}(\cdot)$ at $N=1$ and $M=1$ respectively. ${\mathrm{V}}_1$ is the largest scale BEV feature. The multi-scale stage performs cross-attention across $M$ scales and samples one point per scale. The multi-sample stage uses the output from the multi-scale stage and focuses on the largest scale feature to sample $N$ points. DMD cross-attention reduces the computation complexity to $O(M+N)$ and achieves higher performance than vanilla MSDA.

\section{Experiments}

\begin{table*}[th!]
\centering

\begin{tabular}{c|c|c|c|c|c|c|c|c}
\toprule
Method                & Modality & Backbone & Epoch & AP$_\mathrm{div}$ & AP$_\mathrm{ped}$ & AP$_\mathrm{bou}$ & mAP           & FPS \\

\midrule

% HDMapNet~\pub{ICRA22} & C      & EB0        & 30  & 14.4 & 21.7 & 33.0 & 23.0          & 1.1 \\
% MapTR~\pub{ICLR23}    & C      & R50        & 24  & 51.5 & 46.3 & 53.1 & 50.3          & \textbf{18.0} \\
MapTRv2~\pub{arxiv23} & C      & R50        & 24  & 60.5 & 60.5 & 61.8 & 60.9          & \textbf{16.7} \\
MGMap*~\pub{CVPR24}    & C      & R50        & 24  & 65.0 & 61.8 & 67.5 & 64.8          & - \\
HIMap~\pub{CVPR24}    & C      & R50        & 30  & 68.4 & 62.6 & \textbf{69.1} & 66.7          & - \\
InsMapper~\pub{ECCV24}    & C      & R50        & 24  & 65.1 & 61.6 & 64.6 & 63.8          & - \\
MapQR~\pub{ECCV24}    & C      & R50        & 24  & 68.7 & 63.4 & 67.7 & 66.4          & 16.2 \\
\rowcolor[HTML]{E7E6E6} 
{\mapnet}~(Ours)      & C      & R50        & 24  & \textbf{69.0} & \textbf{64.0} & 68.2 & \textbf{67.1} & 13.9 \\ 
%$^{\ddagger}$
\midrule 

% MapTR~\pub{ICLR23}    & C      & R50        & 110 & 59.8 & 56.2 & 60.1 & 58.7          & \textbf{18.0} \\
MapTRv2~\pub{arxiv23} & C      & R50        & 110 & 68.3 & 68.1 & 69.7 & 68.7          & 16.7 \\
MGMap~\pub{CVPR24} & C      & R50        & 110 & 64.4 & 67.6 & 67.7 & 66.5          & \textbf{18.0} \\
MapQR~\pub{ECCV24}    & C      & R50        & 110  & \textbf{74.4} & 70.1 & 73.2 & 72.6          & 16.2 \\
\rowcolor[HTML]{E7E6E6} 
{\mapnet}~(Ours)      & C      & R50        & 110 &     73.2 &   \textbf{71.5}   &  \textbf{73.3}    & \textbf{72.7}          &  13.9 \\ 

\midrule 

HDMapNet~\pub{ICRA22}        & C \& L & EB0 \& PP  & 30 & 29.6 & 16.3 & 46.7 & 31.0          & 0.6 \\
% MapTR~\pub{ICLR23}    & C \& L & R50 \& Sec & 24  & 62.3 & 55.9 & 69.3 & 62.5          & - \\
MapTRv2~\pub{arxiv23} & C \& L & R50 \& Sec & 24  & 66.5 & 65.6 & 74.8 & 69.0          & \textbf{7.8} \\
MGMap~\pub{CVPR24} & C \& L     & R50 \& Sec   & 24 & 71.1 & 67.7 & \textbf{76.2} & 71.7          & 7.5 \\
\rowcolor[HTML]{E7E6E6} 
{\mapnet}~(Ours)      & C \& L & R50 \& Sec & 24  &   \textbf{72.4}   &  \textbf{70.1}    &   \textbf{76.2}   &         \textbf{72.9}   & 7.5 \\ 

\bottomrule

\end{tabular}
\caption{Quantitative evaluation on nuScenes \texttt{val} set.  ``C'' and ``L'' respectively refer to multi-view cameras and LiDAR inputs. ``R50'', ``EB0'', ``PP'' and ``Sec'' denote  ResNet50~\cite{he2016resnet}, EfficientNet-B0~\cite{tan2019efficientnet}, PointPillars~\cite{lang2019pointpillars} and SECOND~\cite{yan2018second} respectively. ``MGMap*'' means MGMap based on MapTRv2. FPS is tested on a single RTX 4090 GPU for fair comparison.  ``-'' means the corresponding results are not available.}
% {\mapnet} outperforms previous SOTA methods.
% ``${}^\ddagger$'' is reproduced by ourselves.
% The best result is highlighted in \textbf{bold}.
\label{tab:nus_main}
\end{table*}  % \ref{tab:nus_main}
\begin{table}[t!]
\centering
\setlength{\tabcolsep}{1.2mm}
\begin{tabular}{c|c|c|c|c|c|c}
\toprule
Dim. & Method  & AP$_\mathrm{div}$ & AP$_\mathrm{ped}$ & AP$_\mathrm{bou}$ & mAP  & FPS           \\
\midrule
         & MapTRv2 & 71.5              & 63.6              & 67.4              & 67.5 & \textbf{15.2} \\
         & HIMap   & 69.5              & 69.0              & 70.3              & 69.6 & -             \\
         & MapQR   & 72.3              & 64.3              & 68.1              & 68.2 & 14.2          \\
\multirow{-4}{*}{2} &
  \cellcolor[HTML]{E7E6E6}{\mapnet} &
  \cellcolor[HTML]{E7E6E6}\textbf{75.4} &
  \cellcolor[HTML]{E7E6E6}\textbf{69.3} &
  \cellcolor[HTML]{E7E6E6}\textbf{71.3} &
  \cellcolor[HTML]{E7E6E6}\textbf{72.0} &
  \cellcolor[HTML]{E7E6E6}12.6 \\
\midrule
         & MapTRv2 & 68.9              & 60.7              & 64.5              & 64.7 & \textbf{15.0} \\
         & HIMap   & 68.3              & \textbf{66.7}     & \textbf{70.3}     & 68.4 & -             \\
         & MapQR   & 71.2              & 60.1              & 66.2              & 65.9 & 14.1          \\
\multirow{-4}{*}{3} &
  \cellcolor[HTML]{E7E6E6}{\mapnet} &
  \cellcolor[HTML]{E7E6E6}\textbf{73.4} &
  \cellcolor[HTML]{E7E6E6}66.5 &
  \cellcolor[HTML]{E7E6E6}69.8 &
  \cellcolor[HTML]{E7E6E6}\textbf{69.9} &
  \cellcolor[HTML]{E7E6E6}12.6\\

\bottomrule

\end{tabular}
\caption{Quantitative evaluation on Argoverse 2 \texttt{val} set. {\mapnet} reaches SOTA performance, validating its generalization ability. Since Argoverse~2 provides 3D map annotation, ``Dim.'' represents the dimension used to model map elements. When the map dimension is 2, the height information of map elements is dropped, and when the map dimension is 3, the 3D map elements are predicted directly. FPS is tested on a single RTX 4090 GPU for fair comparison. ``-'' means the corresponding results are not available.
}
\label{tab:av2_main}
\end{table}

% Method                & Map Dim. & Backbone & Epoch & AP$_\mathrm{div}$ & AP$_\mathrm{ped}$ & AP$_{bou}$ & mAP           & FPS \\

% \midrule

% % MapTR~\pub{ICLR23}    & 2 & R50 & 6 & 58.1 & 54.7 & 56.7 & 56.5          &  \\
% MapTRv2~\pub{arXiv23} & 2 & R50 & 6 & 71.5 & 63.6 & 67.4 & 67.5          &  \textbf{15.2} \\
% HIMap~\pub{CVPR24}    & 2 & R50 & 6 & 69.5 & 69.0 & 70.3 & 69.6          & - \\
% MapQR~\pub{ECCV24}    & 2 & R50 & 6 & 72.3 & 64.3 & 68.1 & 68.2          & 14.2 \\
% \rowcolor[HTML]{E7E6E6} 
% {\mapnet}~(Ours)      & 2 & R50 & 6 &  \textbf{75.4}    &    \textbf{69.3}  &   \textbf{71.3}   & \textbf{72.0}              & 12.6 \\

% \midrule

% MapTRv2~\pub{arXiv23} & 3 & R50 & 6 & 68.9 & 60.7 & 64.5 & 64.7          & \textbf{15.0} \\
% HIMap~\pub{CVPR24}    & 3 & R50 & 6 & 68.3 & \textbf{66.7} & \textbf{70.3} & 68.4          & - \\
% MapQR~\pub{ECCV24}    & 3 & R50 & 6 & 71.2 & 60.1 & 66.2 & 65.9 & 14.1 \\
% \rowcolor[HTML]{E7E6E6} 
% {\mapnet}~(Ours)      & 3 & R50 & 6 &  \textbf{73.4}   &  66.5   &   69.8  & \textbf{69.9}          & 12.6 \\  % \ref{tab:av2_main}

\subsection{Datasets and Metrics}

To validate the effectiveness of our proposed method {\mapnet}, we evaluate it on the widely used nuScenes dataset~\cite{caesar2020nuscenes} and Argoverse~2 dataset~\cite{wilson2023argoverse} and compare it with the SOTA methods.

The nuScenes dataset is a standard benchmark for online vectorized HD map construction, featuring 1000 driving scenes captured by six multi-view cameras and LiDAR, with 2D vectorized map elements as ground truth. Argoverse~2 is designed for perception and prediction studies in autonomous driving, containing 1000 scenes of 15 seconds each. 3D vectorized map elements captured by seven multi-view cameras are provided as ground truth.

Following the previous studies~\cite{li2022hdmapnet, liao2022maptr}, we evaluate performance across three categories of map elements: lane dividers, pedestrian crossings, and road boundaries. The performance of {\mapnet} is assessed using the Average Precision (AP) metric, where a prediction is considered a True Positive if the Chamfer Distance between the prediction and its ground truth is within thresholds of 0.5, 1.0, and 1.5 meters.

\subsection{Implementation Details}

Our model is trained on $8$ NVIDIA A100 GPUs with a batch size of $8\times3$. Unless otherwise specified, the number of training epochs is 24 on nuScenes and 6 on Argoverse~2. The BEV range is [-30m, 30m] along the longitudinal axis and [-15m, 15m] along the lateral axis, with a feature size ${H_{{\mathrm{}}}}\times{W_{{\mathrm{}}}}$ of 200$\times$100. The number of instance queries $N_I$, prior queries $N_\mathrm{pri}$ and point queries $N_P$ are set to 50, 9 and 20, respectively. Both $\lambda_1$ and $\lambda_2$ are set to 1. $\delta_\mathrm{v}$ is 0.5 and $\delta_\mathrm{d}$ is 3. Other settings keep in line with MapTRv2. The supplementary material shows more implementation details and ablation studies on hyperparameters. 

\subsection{Main Results}
\begin{figure*}[t!]
\centering
\includegraphics[scale=0.111]{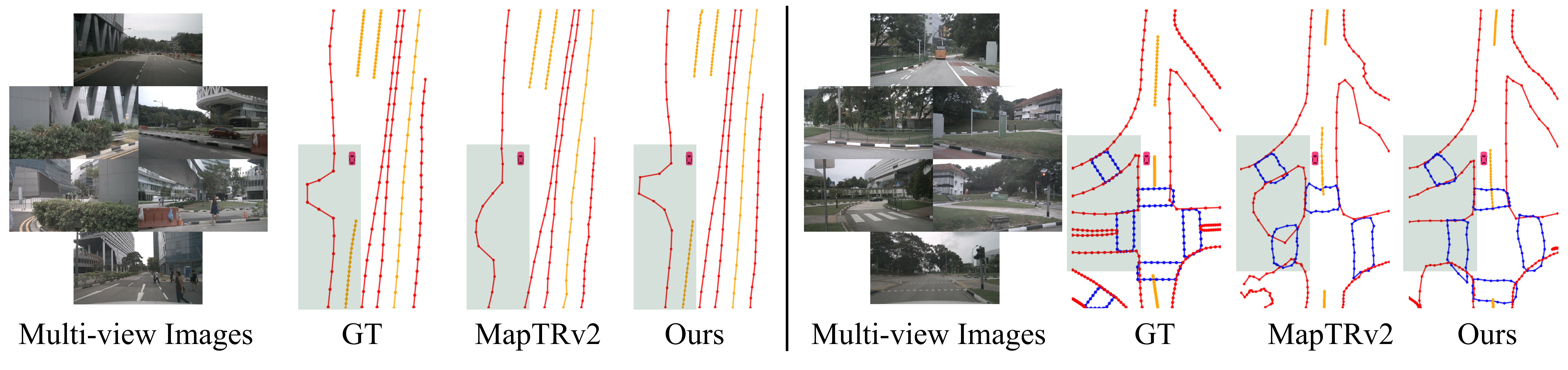}
\caption{Qualitative results on nuScenes \texttt{val} set. We compare visualization results of {\mapnet} with MapTRv2 and corresponding GTs. Models are trained for 24 epochs. The green area indicates that our method achieves more accurate results.}
\label{fig:visualization}
\end{figure*}

\textbf{Results on nuScenes.} We report quantitative results on nuScenes \texttt{val} set in Tab.~\ref{tab:nus_main}. Under camera modality, {\mapnet} surpasses previous SOTA methods and achieves 6.2\% mAP improvement compared with our baseline MapTRv2. On one RTX 4090 GPU, {\mapnet} infers at 13.9 frames per second~(FPS). Additionally, under the camera and lidar fusion modality, {\mapnet} reaches 72.9\% mAP and 7.5 FPS, demonstrating strong generalization capabilities. Qualitative results are shown in Fig.~\ref{fig:visualization}, further illustrating that {\mapnet} achieves improved results. More qualitative results are shown in the supplementary material.

\noindent\textbf{Results on Argoverse 2.} We report quantitative results on Argoverse 2 \texttt{val} set in Tab.~\ref{tab:av2_main}. Argoverse~2 provides 3D map annotations, allowing predictions for both 2D and 3D map elements. {\mapnet} surpasses previous SOTA methods in both dimensions, achieving 72.0\% mAP for 2D map elements and 69.9\% mAP for 3D map elements with an inference speed of 12.6  FPS. Experimental results demonstrate the generalizability of our method.

\noindent\textbf{Results on Enlarged BEV Range.} We train and evaluate models on enlarged BEV ranges on nuScenes \texttt{val} set as shown in Tab.~\ref{tab:largerange}. The size of the BEV grid is maintained at [0.3m, 0.3m]. To verify the robustness of our method, we correspondingly enlarge the prior clustering and position range of map elements. Other settings remain in line with the original models. Experimental results demonstrate that {\mapnet} maintains superiority on enlarged BEV ranges. Notably, under the range of 100 $\times$ 50m, our method outperforms the SOTA method SQD-MapNet~\cite{wang2024sqd} which integrates stream strategy.

\begin{table}[t!]
\centering
\setlength{\tabcolsep}{1.2mm}
\begin{tabular}{c|c|c|c|c|c}
\toprule
Range                             & Method     & AP$_\mathrm{div}$ & AP$_\mathrm{ped}$ & AP$_\mathrm{bou}$ & mAP           \\
\midrule
\multirow{2}{*}{90$\times$30m}  & MapTRv2    & 61.9              & 56.9              & 61.4              & 60.0          \\
                                  & {\mapnet}  & \textbf{66.5}     & \textbf{62.0}     & \textbf{66.1}     & \textbf{64.9} \\
                                  \midrule
\multirow{2}{*}{60$\times$60m}  & MapTRv2    & 61.5              & 58.8              & 61.0              & 60.5          \\
                                  & {\mapnet}  & \textbf{66.6}     & \textbf{63.8}     & \textbf{66.5}     & \textbf{65.6} \\
                                  \midrule
\multirow{3}{*}{100$\times$50m} & MapTRv2    & 62.1              & 55.3              & 61.4              & 59.6          \\
                                  & SQD-MapNet & 65.5              & \textbf{67.0}     & 59.5              & 64.0          \\
                                  & {\mapnet}  & \textbf{67.4}     & 62.5              & \textbf{65.0}     & \textbf{65.0}\\
\bottomrule
\end{tabular}
\caption{Quantitative evaluation on nuScenes dataset with enlarged BEV ranges. {\mapnet} maintains superiority.}
\label{tab:largerange}
\end{table}

\subsection{Ablation Study}
We conduct ablation experiments on nuScenes to verify the effectiveness of our proposed modules and their designs. All models are trained for 24 epochs, and the BEV range is 60$\times$30m. As shown in Tab.~\ref{tab:ablation-main}, starting from MapTRv2 as our baseline, each module improves mAP. In addition, we also report the total unstable matching score $u_t$ on the validation set, demonstrating that the prior position and structure of map elements improve the stability of matching.

\begin{table}[t!]
\centering
\setlength{\tabcolsep}{1.4mm}
\begin{tabular}{c|c|c|c|c|c|c|c}
\toprule
PF         & PPS        & DMD & AP$_\mathrm{div}$ & AP$_\mathrm{ped}$ & AP$_\mathrm{bou}$ & mAP & $u_t$ \\
\midrule
-          & -          & -   & 60.5       & 60.5       & 61.8       & 60.9 & 0.413 \\
\checkmark & -          & -   & 64.9       & 61.8       & 64.9       & 63.9 & - \\
-          & \checkmark & -   & 64.6       & 62.8       & 65.8       & 64.4 & 0.365 \\
\checkmark & \checkmark & -   & 68.7       & 63.6       & 68.0       & 66.8 & - \\
\rowcolor[HTML]{E7E6E6} 
\checkmark & \checkmark & \checkmark & \textbf{69.0} & \textbf{64.0} & \textbf{68.2} & \textbf{67.1} & \textbf{0.362}\\
\bottomrule
\end{tabular}
\caption{Ablation study of our modules on nuScenes dataset. ``PF'', ``PPS'' and ``DMD'' denote PF-Encoder, PPS-Decoder and DMD cross-attention, respectively. $u_t$ is the total unstable matching score defined in Section~\ref{sec:intro}.}
\label{tab:ablation-main}
\end{table}

\noindent\textbf{Ablation study of decoder query prior.} 
We compare various query priors under different BEV feature scales in Tab.~\ref{tab:ablation-decoder}. MAI in MGMap provides semantic priors for queries, but the improvement is limited. This limitation arises because semantic priors lack positional information crucial for vectorized map elements. We then integrate MAI with query position, but the mAP decreases as the semantic priors are not well-suited for query position. In contrast, our PPS provides prior position and structure, resulting in significant performance improvements. Compared with directly using clustering results as priors, abstracted priors perform better.

\noindent\textbf{Ablation study of PF-Encoder design.}
We compare various BEV feature priors and encoders, as shown in Tab.~\ref{tab:ablation-encoder}. All GKT and BEVFormer methods utilize a single-layer architecture. To ensure a fair comparison, we incorporate auxiliary depth supervision across all experimental setups. LSS without prior is the encoder of MapTRv2. The third row denotes a two-layer BEVFormer encoder to exclude the impact of the number of encoder layers. The results show that LSS serves as a good prior generator. Results also demonstrate the effectiveness of Discriminative Loss.

\noindent\textbf{Ablation study of DMD cross-attention design.} We compare different decoupling designs of DMD cross-attention, including parallel and serial connection of two MSDA mechanisms, as shown in Tab.~\ref{tab:ablation-dmd}. In the serial connection, we compare different performance orders of MSDA along multi-sample and multi-scale dimensions. Experimental results demonstrate that our proposed DMD cross-attention improves efficiency and achieves superior performance. 

\begin{table}[t!]
\centering
\setlength{\tabcolsep}{1.5mm}
\begin{tabular}{c|c|c|c|c|c}
\toprule
${F}_{\mathrm{BEV}}$ & Query Prior & AP$_\mathrm{div}$    & AP$_\mathrm{ped}$    & AP$_\mathrm{bou}$    & mAP           \\
\midrule
                             & -           & 64.9          & 61.8          & 64.9          & 63.9          \\
                             & MAI         & 65.0          & 62.7          & 64.8          & 64.2          \\
                             & MAI-pos     & 58.3          & 51.0          & 62.3          & 57.2          \\
                             & PPS-clst     & 64.6          & \textbf{63.8}          & 66.9          & 65.1          \\
\multirow{-4}{*}{\begin{tabular}[c]{@{}c@{}}Single\\ Scale\end{tabular}} &
  PPS &
  \textbf{65.5} &
  62.8 &
  \textbf{68.5} &
  \textbf{65.6} \\
\midrule
                             & -           & 67.5          & 61.7          & 65.8          & 65.0          \\
                             & MAI         & 66.5         & 62.1          & 67.1          & 65.2          \\
                             & MAI-pos     & 58.4          & 54.7          & 61.2          & 58.1          \\
                             & PPS-clst     & 68.4          & 63.3          & 67.5          & 66.4          \\
\multirow{-4}{*}{\begin{tabular}[c]{@{}c@{}}Multi\\ Scales\end{tabular}} &
  \cellcolor[HTML]{E7E6E6}PPS &
  \cellcolor[HTML]{E7E6E6}\textbf{68.7} &
  \cellcolor[HTML]{E7E6E6}\textbf{63.6} &
  \cellcolor[HTML]{E7E6E6}\textbf{68.0} &
  \cellcolor[HTML]{E7E6E6}\textbf{66.8}\\
\bottomrule
\end{tabular}
\caption{Ablation study of decoder query prior. ``MAI'' denotes Mask-Activated Instance in MGMap, and ``MAI-pos'' denotes integrating MAI with query position. ``PPS-clst'' denotes directly using clustering results as priors.}
\label{tab:ablation-decoder}
\end{table}

\begin{table}[t!]
\centering
\setlength{\tabcolsep}{1.5mm}
\begin{tabular}{c|c|c|c}
\toprule
${F}_{\mathrm{BEV}}$ Prior & BEV Encoder & Dis. Loss  & mAP           \\
\midrule
-                   & LSS-d       & -          & 60.9          \\
-                   & BEVFormer-d & -          & 57.3          \\
\midrule
BEVFormer-d               & BEVFormer   & -          & 59.6          \\
GKT-d               & BEVFormer   & -          & 61.9          \\
LSS-d               & GKT         & -          & 62.4          \\
LSS-d               & BEVFormer   & -          & 63.3          \\
\rowcolor[HTML]{E7E6E6} 
LSS-d               & BEVFormer   & \checkmark & \textbf{63.9}\\
\bottomrule
\end{tabular}
\caption{Ablation study of the PF-Encoder design. 
% We compare different BEV feature priors and encoders. 
``Dis. Loss'' denotes  Discriminative Loss, and ``-d'' denotes auxiliary depth supervision of image features.}
\label{tab:ablation-encoder}
\end{table}

\begin{table}[t!]
\setlength{\tabcolsep}{0.9mm}
\begin{tabular}{c|c|c|c|c|c}
\toprule
MSDA              & AP$_\mathrm{div}$ & AP$_\mathrm{ped}$ & AP$_\mathrm{bou}$ & mAP           & $t_c$~(ms)          \\
\midrule
Vanilla           & 68.7              & 63.6              & 68.0              & 66.8          & 6.1          \\
Parallel Connection & 68.5              & \textbf{65.7}     & 66.5              & 66.9          & 4.2 \\
Sample-then-Scale & 67.6              & 64.7     & 68.0              & 66.8          & \textbf{4.2} \\
\rowcolor[HTML]{E7E6E6} 
Scale-then-Sample & \textbf{69.0}     & 64.0              & \textbf{68.2}     & \textbf{67.1} & \textbf{4.2}\\
\bottomrule
\end{tabular}
\caption{Ablation study of the DMD cross-attention design. ``Sample-then-Scale'' denotes performing MSDA along multi-sample then multi-scale, and ``Scale-then-Sample'' denotes performing MSDA along reverse order. $t_c$ denotes the inference time of cross-attention.}
\label{tab:ablation-dmd}
\end{table}

\section{Conclusion}
In this paper, we introduce {\mapnet} to enhance online vectorized HD map construction with priors. 
To address the issue of unstable matching, we propose the PPS-Decoder, which provides reference points with position and structure priors clustered from the dataset. 
To embed BEV features effectively, we propose the PF-Encoder which enhances the image-to-BEV transformation with BEV feature priors and leverages Discriminative Loss to improve the aggregation of map element embeddings. 
To reduce the computation complexity, we propose the DMD cross-attention, which performs cross-attention respectively along multi-scale and multi-sample.
Our proposed {\mapnet} achieves state-of-the-art performance on nuScenes and Argoverse2 datasets. 

\noindent\textbf{Limitations and Future Work.}
Despite our development for online vectorized HD map construction, several limitations need to be addressed in future work. Firstly, our map element priors only incorporate positional information and lack semantic information, which limits the interaction and optimization of queries. Secondly, our method relies solely on single-frame sensor input, constraining the representation of temporally and spatially continuous map elements.

%

%-------------------------------------------------------------------------
%-------------------------------------------------------------------------

\bibliography{prior}

% supplementary
\newpage
\twocolumn[
\begin{center}
    \LARGE \textbf{Supplementary Material}
\end{center}
\vspace{0.5cm}
]

\setcounter{table}{0}   %从0开始编号，显示出来表会A1开始编号
\setcounter{figure}{0}
\setcounter{section}{0}
\setcounter{equation}{0}

\renewcommand{\thetable}{S\arabic{table}}
\renewcommand{\thefigure}{S\arabic{figure}}
\renewcommand{\thesection}{S\arabic{section}}
\renewcommand{\theequation}{S\arabic{equation}}

\begin{figure*}[t!]
	\centering
        \subfloat[MapTRv2 Decoder
    	\label{fig:decoder-maptrv2-supp}
        ]{
	\begin{minipage}{0.31\linewidth}{
			\begin{center}
            {\includegraphics[width=1\linewidth]{figures/decoder-maptrv2.pdf}}
			\end{center}
		}
	\end{minipage}
        }
        \subfloat[MGMap Decoder
    	\label{fig:decoder-mgmap-supp}
        ]{
	\begin{minipage}{0.32\linewidth}{
			\begin{center}
            {\includegraphics[width=1\linewidth]{figures/decoder-mgmap.pdf}}
			\end{center}
		}
	\end{minipage}
        }
        \subfloat[HIMap Decoder
            \label{fig:decoder-himap-supp}
        ]{
        \begin{minipage}{0.32\linewidth}{
			\begin{center}
			{\includegraphics[width=1\linewidth]{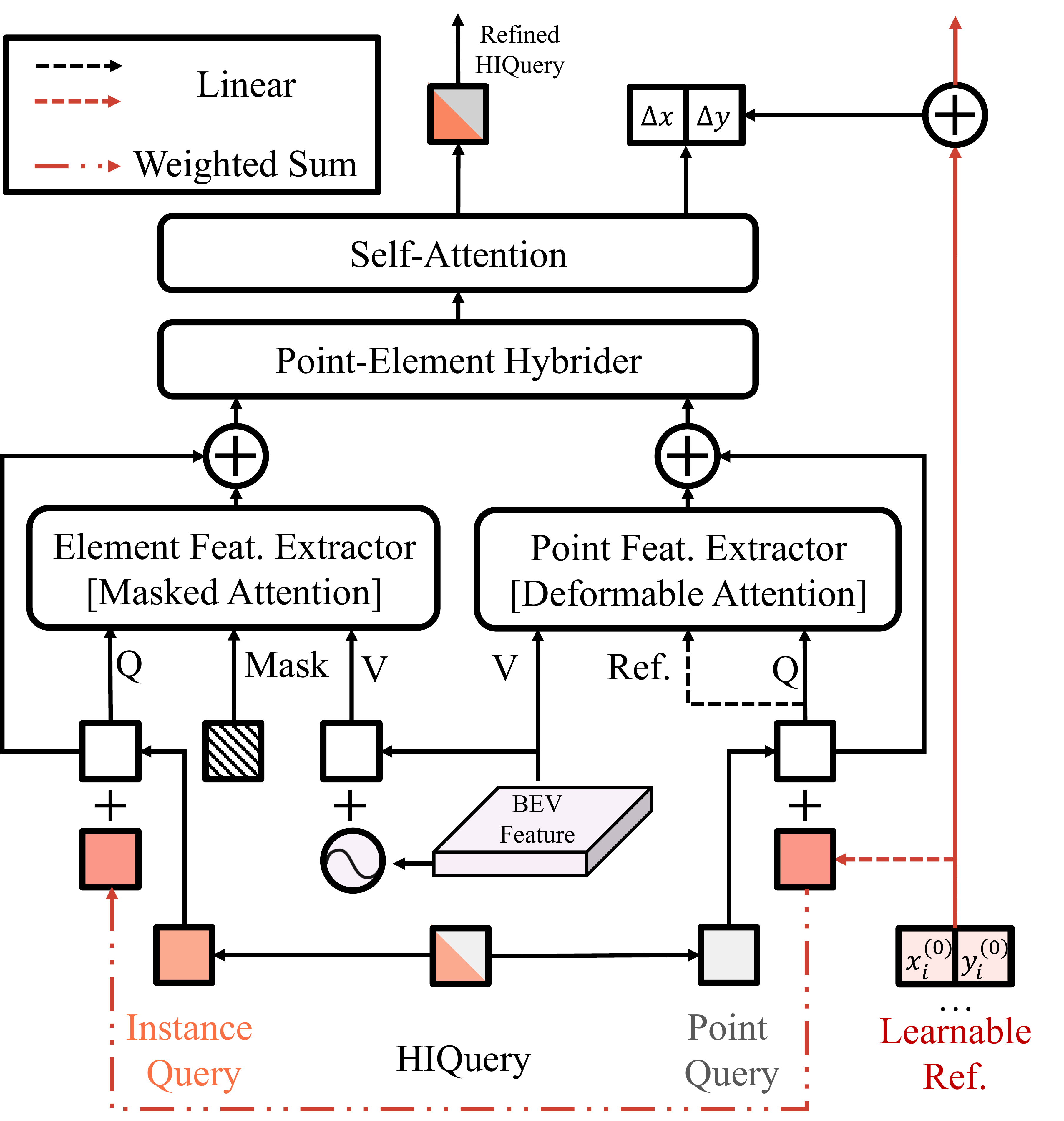}}
			\end{center}
		}
	\end{minipage}
        }
        \\
        \subfloat[MapQR Decoder
    	\label{fig:decoder-mapqr-supp}
        ]{
	\begin{minipage}{0.3\linewidth}{
			\begin{center}
            {\includegraphics[width=1\linewidth]{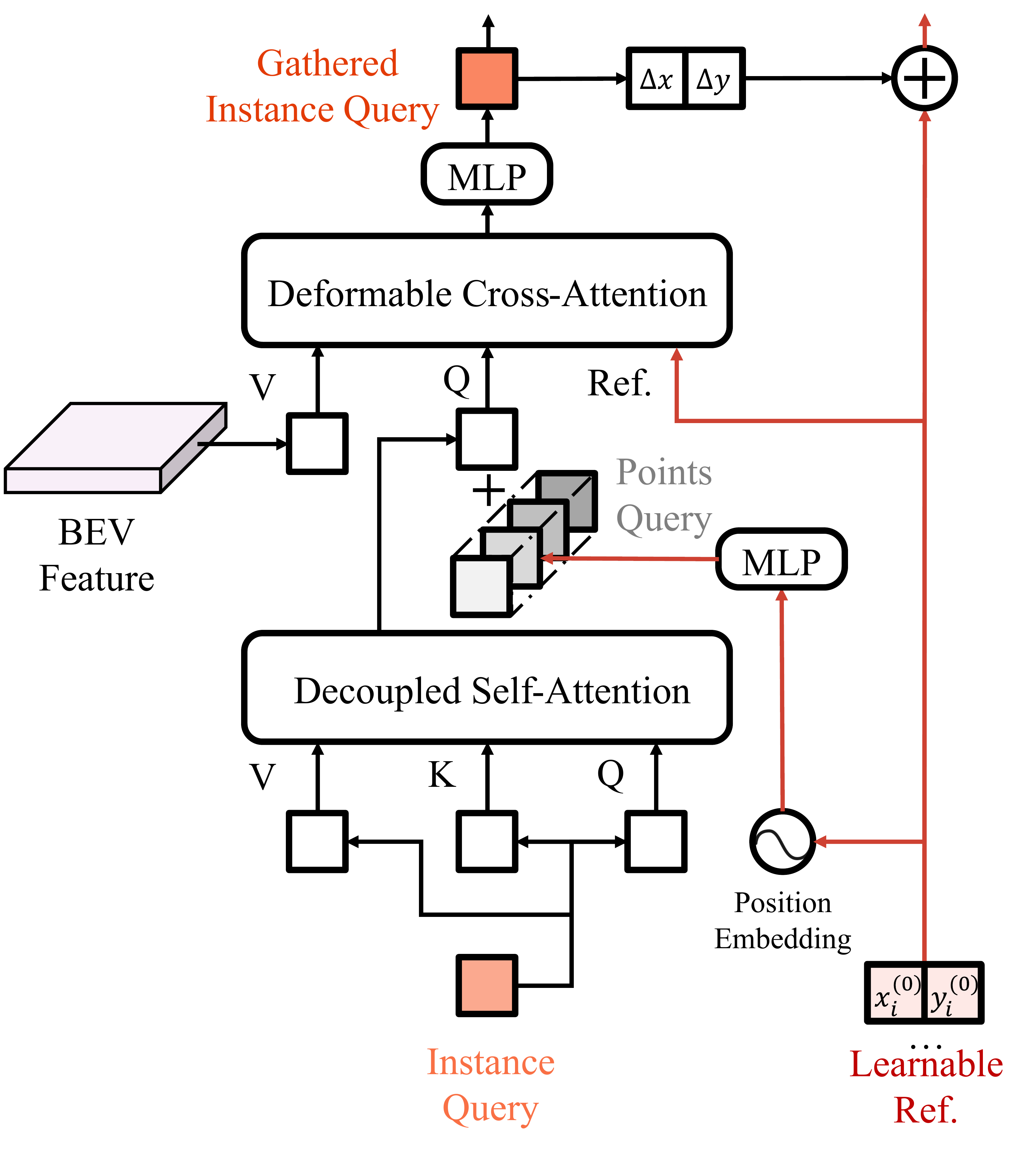}}
			\end{center}
		}
	\end{minipage}
        }
        \hspace{4mm}
        \subfloat[InsMapper Decoder
            \label{fig:decoder-insmapper-supp}
        ]{
        \begin{minipage}{0.235\linewidth}{
			\begin{center}
			{\includegraphics[width=1\linewidth]{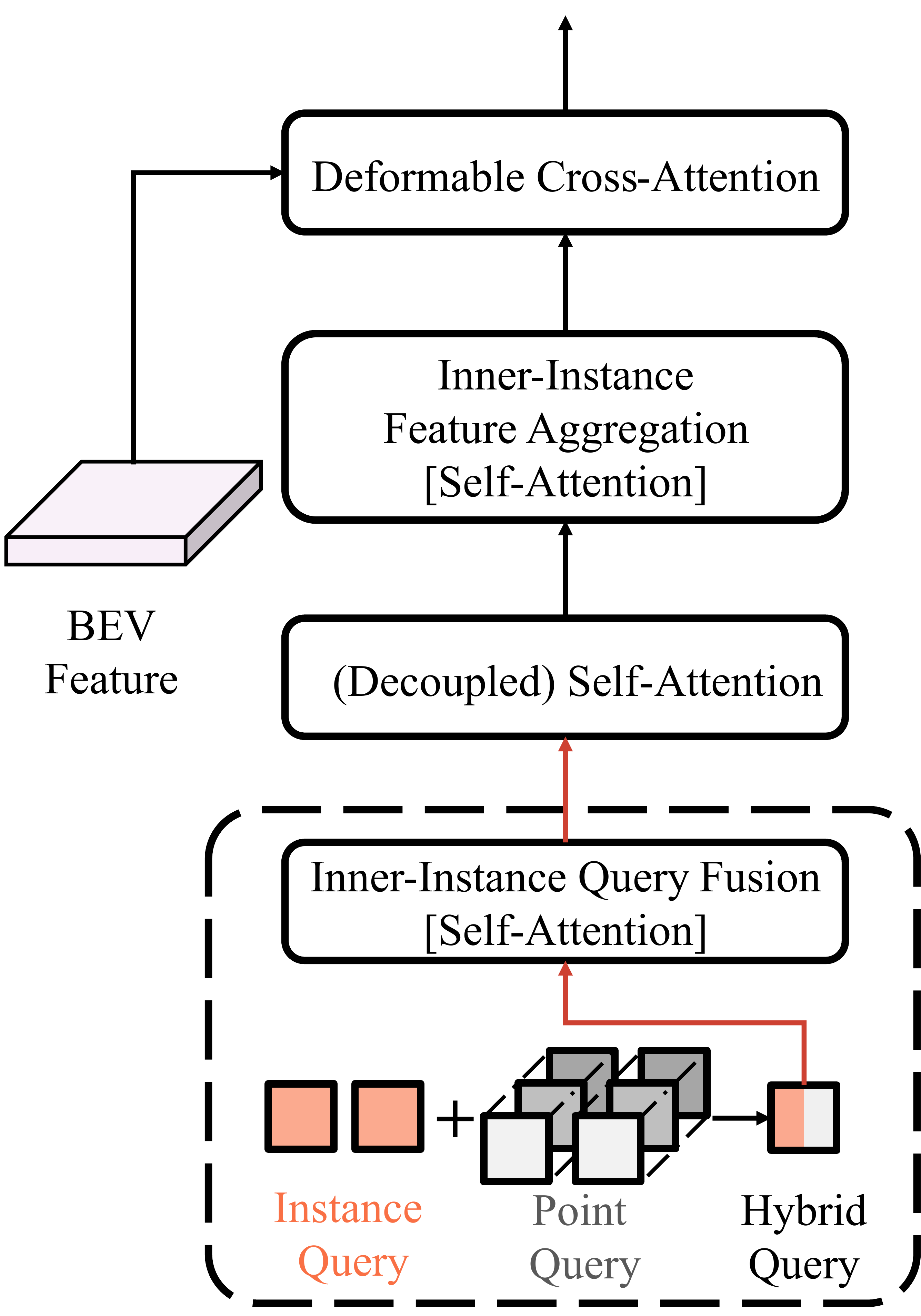}}
			\end{center}
		}
	\end{minipage}
        }
        \hspace{5mm}
        \subfloat[Our PPS-Decoder
            \label{fig:decoder-proposed-supp}
        ]{
        \begin{minipage}{0.32\linewidth}{
			\begin{center}
			{\includegraphics[width=1\linewidth]{figures/decoder.pdf}}
			\end{center}
		}
	\end{minipage}
        }
	\caption{Comparison of the decoders of previous SOTA methods: (a) MapTRv2, (b) MGMap, (c) HIMap, (d) MapQR, (e) InsMapper and (f) our proposed {\mapnet}. (a), (b) and (d) are based on their open-source code, and (c) is based on the implementation details and equations from the paper. For simplicity, we only show the first layer in the transformer decoder. Since InsMapper currently has no open-source code and the specific implementation process is not detailed in the paper, we only show the general structure of its decoder. To highlight the key distinctions between our PriorMapNet and previous SOTA methods, we mark the randomly initialized learnable query position embeddings and reference points in light red, and mark our prior-aware query position embeddings and reference points in yellow.}
\label{fig:decoder-supp}
\end{figure*}

\noindent In the supplementary material, we provide more details, experiments, and analysis of the proposed {\mapnet}, including:
\begin{itemize}
    \item More implementation details of our method;
    \item Additional ablation studies on hyperparameters;
    \item Difference with previous SOTA methods;
    \item More qualitative results and failure cases.
\end{itemize}

\section{More Implementation Details}
\noindent\textbf{Feature extraction and multi-modality fusion.} Multi-modality inputs consist of multi-view cameras and LiDAR data.
For backbones, we utilize ResNet50~\cite{he2016resnet} for image features extraction and SECOND~\cite{yan2018second} for point cloud features extraction. The input image size is 480$\times$800. The voxel size for point cloud is set to [0.1m, 0.1m, 0.2m], and the LiDAR BEV features are upsampled to align with camera BEV features. For BEV feature fusion, we concatenate camera and LiDAR BEV features along feature channels and use a convolution layer to fuse them. 

\noindent\textbf{Encoder and decoder.} 
Following MapTRv2~\cite{liao2023maptrv2}, we add an auxiliary depth loss for LSS to generate BEV query priors effectively. The PPS-Decoder contains 6 layers to refine the outputs iteratively. In addition to 50 instance-level queries, there are another 300 one-to-many queries that are used to speed up convergence during training. Priors are not used for one-to-many queries.

\noindent\textbf{Training.}
We use the AdamW optimizer~\cite{loshchilov2017decoupled} with an initial learning rate of $6 \times 10^{-4}$, and apply the Cosine Annealing learning rate scheduler with a linear warm-up phase~\cite{loshchilov2016sgdr}. Except for our Discriminative Loss, other training and loss settings keep in line with our baseline MapTRv2.

\section{Additional Ablation Studies}
\noindent\textbf{Abalation study on the number of prior queries.} We cluster 50 map elements and abstract the position information of the first $N_\mathrm{pri}$ elements, as shown in Fig.~\ref{fig:matching}. To most effectively utilize the prior position and structure, we compare the results of different numbers of prior queries (\textit{i.e.} $N_\mathrm{pri}$), as shown in Tab.~\ref{tab:ablation-npri}. Utilizing 9 prior queries achieves the best performance. Too few prior queries cannot fully utilize the prior, while too many prior queries limit the learning ability of learnable queries.

\begin{table}[t!]
\centering
\begin{tabular}{c|c|c|c|c}
\toprule
$N_\mathrm{pri}$ & AP$_\mathrm{div}$ & AP$_\mathrm{ped}$ & AP$_\mathrm{bou}$ & mAP           \\
\midrule
0                & 67.5              & 61.7              & 65.8              & 65.0          \\
5                & 67.2              & 62.7              & 67.8              & 65.9          \\
\rowcolor[HTML]{E7E6E6} 
9                & \textbf{68.7}     & 63.6              & 68.0              & \textbf{66.8} \\
10               & 67.3              & \textbf{63.8}     & \textbf{68.8}     & 66.7          \\
20               & 68.1              & 63.5              & 68.0              & 66.5          \\
50               & 68.4              & 63.3              & 67.5              & 66.4         \\
\bottomrule
\end{tabular}
\caption{Ablation study of the number of prior queries. Utilizing 9 prior queries achieves the best performance.}
\label{tab:ablation-npri}
\end{table}

\noindent\textbf{Abalation study on the loss weights.} The Discriminative Loss is defined as $L_\mathrm{dis}=\lambda_1 L_\mathrm{var}+\lambda_2 L_\mathrm{dist}$. Following LaneNet~\cite{neven2018lanenet}, we set $\lambda_1=\lambda_2=\lambda$. In order to obtain the most effective auxiliary discriminative supervision, we compare the results under different loss weights (\textit{i.e.} $\lambda$), as shown in Tab.~\ref{tab:ablation-lambda}. Experimental results show the effectiveness of Discriminative Loss, and the best performance can be achieved when $\lambda=1$.

\begin{table}[t!]
\centering
\begin{tabular}{c|c|c|c|c}
\toprule
$\lambda$ & AP$_\mathrm{div}$ & AP$_\mathrm{ped}$ & AP$_\mathrm{bou}$ & mAP           \\
\midrule
0       & 64.3     & 60.9              & 64.8              & 63.3          \\
0.5       & \textbf{65.6}     & 61.6              & 63.7              & 63.6          \\
\rowcolor[HTML]{E7E6E6} 
1         & 64.9              & \textbf{61.8}     & \textbf{64.9}     & \textbf{63.9} \\
2         & 65.1              & 60.7              & 64.6              & 63.5          \\
3         & 64.5              & 60.1              & 65.0              & 63.2  \\
\bottomrule
\end{tabular}
\caption{Ablation study of the weight of Discriminative Loss. We set $\lambda_1=\lambda_2=\lambda$ and setting $\lambda=1$ achieves the best performance.}
\label{tab:ablation-lambda}
\end{table}

\noindent\textbf{Abalation study on the borders of the variance and distance loss.} $\delta_\mathrm{v}$ and $\delta_\mathrm{d}$ are respectively the borders of the variance and distance loss. Following LaneNet~\cite{neven2018lanenet}, we set $\delta_\mathrm{d}=6\delta_\mathrm{v}$. To embed the BEV features effectively, we compare the results under different borders of the distance loss (\textit{i.e.} $\delta_\mathrm{d}$), as shown in Tab.~\ref{tab:ablation-delta}. Experiments are conducted under $\lambda=1$. Experimental results show that the best performance can be achieved when $\delta_\mathrm{d}=3$.

\begin{table}[t!]
\centering
\begin{tabular}{c|c|c|c|c}
\toprule
$\delta_\mathrm{d}$ & AP$_\mathrm{div}$ & AP$_\mathrm{ped}$ & AP$_\mathrm{bou}$ & mAP           \\
\midrule
1                   & 64.8              & 60.5              & 64.2              & 63.2          \\
\rowcolor[HTML]{E7E6E6} 
3                   & 64.9              & \textbf{61.8}     & \textbf{64.9}     & \textbf{63.9} \\
6                   & \textbf{66.0}     & 61.2              & 63.9              & 63.7          \\
12                  & 65.4              & 61.0              & 64.5              & 63.6         \\
\bottomrule
\end{tabular}
\caption{Ablation study of the borders of the variance and distance loss. We set $\delta_\mathrm{d}=6\delta_\mathrm{v}$ and setting $\delta_\mathrm{d}=3$ achieves the best performance.}
\label{tab:ablation-delta}
\end{table}

\section{Difference with Previous SOTA Methods}
We highlight the key distinctions between our {\mapnet} and previous SOTA methods, including MapTRv2~\cite{liao2023maptrv2},  MGMap~\cite{liu2024mgmap}, HIMap~\cite{zhou2024himap}, MapQR~\cite{liu2024mapqr} and InsMapper~\cite{xu2023InsMapper}. We compare the decoders in Fig.~\ref{fig:decoder-supp}, which are the core of these methods.

\noindent \textbf{Motivation and model design.} 
Our motivation and model design diverge from those of previous SOTA methods. {\mapnet} starts from the issue of \textit{unstable matching} and utilizes pre-computed clustered priors to initialize reference points. Fitted from the map elements in the dataset, prior reference points lower the learning difficulty and achieve stable matching. 
% Furthermore, we consider the priors to be a more efficient initialization and enhance the encoder with prior features. 
In contrast, HIMap, MapQR, and InsMapper ignore query priors and use randomly initialized reference points. Based on MapTR series methods, they explore the correlations between instances and points, which introduce additional modules and computational complexity. MGMap proposes Mask-Actived Instance to learn map instance segmentation results and provides semantic priors for instance queries. However, semantic priors lack positional information, which is essential for vectorized map elements. Our PPS-Decoder provides prior position and structure, resulting in significant performance improvements.

\noindent \textbf{Performance.} 
Our {\mapnet} surpasses previous SOTA methods on nuScenes and Argoverse~2 dataset, as illustrated in Tab.~\ref{tab:nus_main} and Tab.~\ref{tab:av2_main}. {\mapnet} achieves SOTA performance under different modalities on nuScenes dataset and different dimensions on Argoverse~2 dataset, demonstrating the generalizability of our method.

\section{Qualitative Results and Failure Cases}
\begin{figure*}[t!]
\centering
\includegraphics[scale=0.167]{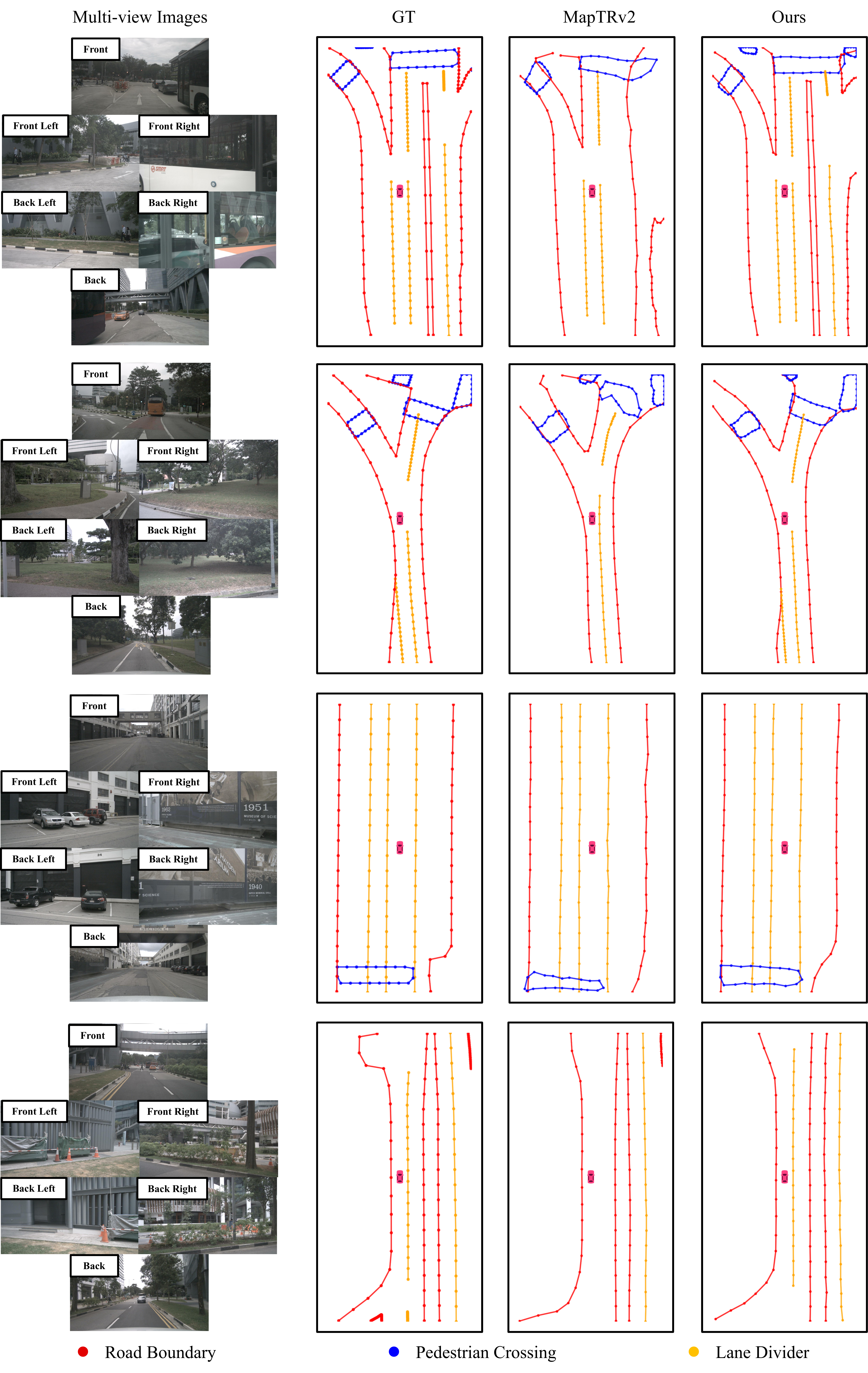}
\caption{Visualization of MapTRv2 and {\mapnet} results and the corresponding GTs. Models are trained for 24 epochs.}
\label{fig:visualization-supp}
\end{figure*}

\begin{figure*}[t!]
\centering
\includegraphics[scale=0.167]{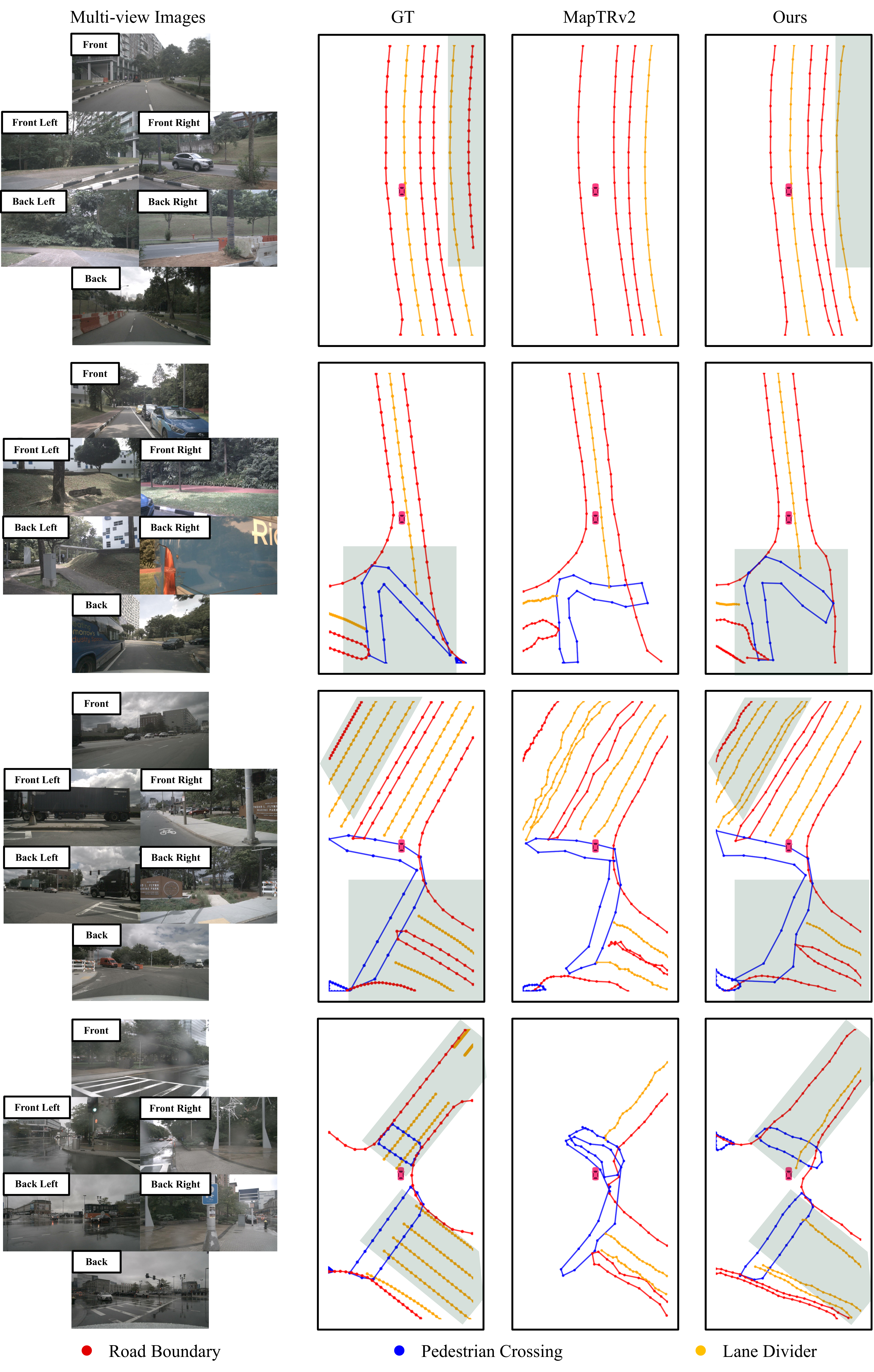}
\caption{Visualization of failure cases. The green areas indicate locations where our method fails to predict accurate results.}
\label{fig:visualization-failure}
\end{figure*}

We show more qualitative results and failure cases on nuScenes \texttt{val} set in Fig.~\ref{fig:visualization-supp} and Fig.~\ref{fig:visualization-failure}, respectively. For each example, the first column represents multi-view camera images. The second column shows the prediction GT. The third and fourth columns respectively show the predicted results of MapTRv2 and our {\mapnet}. Visualization results illustrate that PriorMapNet achieves improved results than MapTRv2. However, in some complex scenes such as intersections, our method fails to predict some map elements, showing our limitations and requiring future work.

\end{document}